\documentclass{article}

\usepackage{preprint}
\AtBeginDocument{%
  \setlength{\headheight}{22.76004pt}%
  \addtolength{\topmargin}{-9.75198pt}%
}
\usepackage{algorithm}
\usepackage{algorithmic}
\usepackage{amsmath, amsthm, amssymb, amsfonts}

\usepackage[numbers,square]{natbib}
\usepackage{xcolor}		
\usepackage[colorlinks = true,
            linkcolor = purple,
            urlcolor  = blue,
            citecolor = cyan,
            anchorcolor = black]{hyperref}	
\usepackage{booktabs} 		
\usepackage{nicefrac}		
\usepackage{microtype}		
\usepackage{lineno}		
\usepackage{float}
\usepackage{multirow}
\usepackage{graphicx}
\usepackage{colortbl}
\usepackage{subcaption}
\usepackage{enumitem}

\usepackage{newfloat}
\DeclareFloatingEnvironment[name={Supplementary Figure}]{suppfigure}
\usepackage{sidecap}
\sidecaptionvpos{figure}{c}

\usepackage{titlesec}
\titlespacing\section{0pt}{12pt plus 3pt minus 3pt}{1pt plus 1pt minus 1pt}
\titlespacing\subsection{0pt}{10pt plus 3pt minus 3pt}{1pt plus 1pt minus 1pt}
\titlespacing\subsubsection{0pt}{8pt plus 3pt minus 3pt}{1pt plus 1pt minus 1pt}

\definecolor{lightgray}{gray}{0.9}
\definecolor{lightgrey}{gray}{0.92}

\title{RobustSora: A De-Watermarked Benchmark for Evaluating Watermark Robustness in AI-Generated Video Detection}

\usepackage{eso-pic}
\usepackage{tikz}
\PassOptionsToPackage{colorlinks=true,linkcolor=gray,urlcolor=gray}{hyperref}

\usepackage{titling}
\usepackage{orcidlink}
\usepackage{footmisc}
\newcommand{\Author}[3]{
  \textbf{#1}\textsuperscript{#2},\ \orcidlink{#3} %
}

\author{
  \Author{Zhuo Wang}{1}{0009-0008-6447-4332} \and
  \Author{Xiliang Liu}{1}{0000-0000-0000-0000} \and
  \Author{Ligang Sun}{1}{0009-0006-5712-5068}
}

\date{%
  \textsuperscript{1}College of Computer Science, Beijing University of Technology, China\\[1em]
  \footnotesize \textbf{Corresponding author:} Xiliang Liu \texttt{<liuxl@bjut.edu.cn>}\\
}

\begin{document}

\maketitle
\thispagestyle{empty}

\begin{abstract}
The proliferation of AI-generated video models poses new challenges to information integrity and digital trust. A key confound, however, remains unaddressed: commercial generators embed visible overlay watermarks for provenance tracking, yet no existing benchmark controls for this variable, leaving open whether detectors learn genuine generation artefacts or merely associate watermark patterns with AI-generated labels. We present \textbf{RobustSora}, a benchmark of 6{,}500 manually verified videos in four categories: Authentic-Clean (A-C), Generated-Watermarked (G-W), Generated-DeWatermarked (G-DeW), and Authentic-Spoofed (A-S), sourced from Vript, DVF, and UltraVideo (authentic) and from Sora, Sora~2, Pika, Open-Sora~2, and KLing (generated). Two evaluation tasks isolate watermark effects: Task-I (Watermark Erasure Robustness) tests detection on watermark-removed AI videos; Task-II (Watermark Spoofing Robustness) measures false-alarm rates on authentic videos injected with fake watermarks. Across ten models spanning specialized detectors, transformer classifiers, and MLLMs, watermark manipulation induces accuracy changes of $-9.4$ to $+1.6$\,pp (mean 6.6\,pp; $p{<}0.01$ for 7/10 models on each task). A placebo control bounds inpainting-artefact confounds at $\le$2\,pp, and a watermark-aware training augmentation recovers 3--4\,pp on both tasks, together providing causal evidence that detectors actively rely on watermark cues. Per-generator breakdown shows that Sora~2 induces drops of $-11$ to $-14$\,pp versus $-3$ to $-6$\,pp for Pika and Open-Sora~2, indicating that watermark prominence, rather than detector architecture, is the principal driver of dependency. These results argue for watermark-aware evaluation and training in AIGC video detection. Dataset, evaluation code, and pretrained checkpoints will be released.
\end{abstract}

\noindent\textbf{Keywords:} AI-generated video detection, digital watermarking, benchmark, robustness evaluation, video forensics, deepfake detection

\vspace{0.35cm}


\section{Introduction}
\label{sec:introduction}

The rapid evolution of video generation technologies has fundamentally transformed the landscape of digital content creation. State-of-the-art models such as Sora~\cite{sora}, Sora~2~\cite{sora2}, KLing~\cite{kling2024_website}, Pika~\cite{pika2024_website}, and Open-Sora~2~\cite{pengOpenSora20Training2025} now produce videos of remarkable visual fidelity and temporal coherence, often approaching the perceptual quality of authentic footage. While these advances unlock tremendous creative potential in filmmaking, education, and entertainment~\cite{zhang2025generative_arxiv2025}, they simultaneously pose severe threats to information integrity, public trust, and media authenticity~\cite{vaccari2020deepfakes_sms2020, hwang2021effects_cyberpsychology2021, zhouSoraIncredibleScary2024}. The potential for weaponized synthetic media---including political deepfakes, financial fraud, and targeted disinformation campaigns---has elevated AIGC video detection from a niche research topic to a critical societal challenge.

In response, the research community has made significant strides in developing detection benchmarks and methods. GenVidBench~\cite{ni2025genvidbench_arxiv2025} established a large-scale evaluation framework comprising 143,000 videos with cross-generator protocols, achieving 79.90\% accuracy with MViT~V2 on challenging test sets. AEGIS~\cite{AEGIS} introduced a hyper-realistic benchmark featuring over 10,000 videos from cutting-edge commercial generators, revealing that even advanced multimodal large language models (MLLMs) such as Qwen2.5-VL achieve only 22--23\% zero-shot accuracy on the most challenging subsets. DuB3D~\cite{jiDistinguishAnyFake2024} leveraged a massive dataset of 2.66 million videos with a dual-branch 3D transformer architecture, achieving 96.77\% in-domain accuracy through the fusion of raw spatio-temporal data and optical flow features. More recently, BusterX++~\cite{wenBusterXUnifiedCrossModal2025} pioneered cross-modal detection and explanation through reinforcement learning post-training, while IVY-FAKE~\cite{zhangIVYFAKEUnifiedExplainable2025} proposed a unified explainable detection framework for both images and videos. On the methodology front, approaches have diversified from spatial artifact-based classifiers~\cite{wang2020cnn, ojha2023towards} to temporal consistency analysis~\cite{maDetectingAIGeneratedVideo2025a, zheng2021exploring}, physics-driven spatiotemporal modeling~\cite{zhangPhysicsDrivenSpatiotemporalModeling2025}, and MLLM-powered reasoning~\cite{DBLP:conf/nips/SongGZL00Z024, wenBusterXUnifiedCrossModal2025}.

Despite this progress, existing benchmarks and detection methods share a critical blind spot: \textit{they do not account for the role of digital watermarks embedded by generative model providers}. Major commercial platforms, notably Sora~2~\cite{sora2}, embed visible or semi-transparent overlay watermarks in their generated outputs as a provenance tracking mechanism. These watermarks, designed for content attribution and policy compliance, introduce spatially localised and temporally consistent patterns into the generated videos. A fundamental question thus arises:

\vspace{0.5em}
\noindent\textit{Do state-of-the-art AIGC video detectors genuinely learn generation artifacts, or do they partially rely on watermark patterns as a shortcut for classification?}
\vspace{0.5em}

This question has implications for the real-world deployment of detection systems. If detectors learn to associate watermark patterns with ``AI-generated'' labels, they become vulnerable to two classes of adversarial attacks: (1)~\textbf{Watermark erasure attacks}, where adversaries remove watermarks from AI-generated videos to evade detection, and (2)~\textbf{Watermark spoofing attacks}, where adversaries inject fake watermarks into authentic videos to trigger false positives. Both attack vectors undermine the reliability of detection systems in practical deployment scenarios.

The relevance of this concern is underscored by concurrent research on watermark fragility. VideoMarkBench~\cite{jiang2025videomarkbenchbenchmarkingrobustnessvideo} recently demonstrated that existing video watermarking methods---including REVMark\cite{zhang2023novel}, StegaStamp\cite{tancik2020stegastamp}, VideoSeal\cite{fernandez2024video}, and VideoShield\cite{huVideoShieldRegulatingDiffusionbased2025}---are systematically vulnerable to both removal and forgery perturbations across white-box, black-box, and no-box threat models. Their findings reveal that all evaluated watermarking methods fail completely under white-box attacks, and remain vulnerable to adversarial removal in black-box settings. This fragility of watermarks has a direct implication for detection: if watermarks can be reliably removed or forged, then any detection system that depends on watermark cues is fundamentally compromised.

To systematically investigate the influence of watermarks on AIGC video detection, we propose \textbf{RobustSora}, a comprehensive benchmark designed to quantify watermark dependency in current detection paradigms. Our benchmark makes the following key contributions:

\begin{enumerate}[leftmargin=*]
\item \textbf{The first watermark robustness benchmark for AIGC video detection.} RobustSora is the first benchmark explicitly designed to evaluate how watermark presence and absence affect detector performance, addressing a critical gap in the evaluation landscape.

\item \textbf{A systematically constructed four-type dataset.} We construct a dataset of 6,500 high-quality videos across four categories---Authentic-Clean (A-C), Authentic-Spoofed (A-S), Generated-Watermarked (G-W), and Generated-DeWatermarked (G-DeW)---sourced from diverse authentic and generative origins (see Figure~\ref{fig:overview}).

\item \textbf{A dual-task evaluation protocol.} We introduce two complementary evaluation tasks: Task-I quantifies the impact of watermark removal on detection of AI-generated videos, while Task-II assesses the susceptibility of detectors to watermark-based spoofing attacks on authentic content.

\item \textbf{Comprehensive cross-architecture analysis.} We benchmark ten state-of-the-art models spanning three architectural paradigms---specialized AIGC detectors (DeCoF, NSG-VD, D3), transformer-based classifiers (TimeSformer, VideoSwin-T, MViT~V2, DuB3D-FF), and MLLM-based approaches (Qwen2.5-VL-3B/7B, Video-LLaVA-7B)---revealing architecture-specific watermark dependency patterns.

\item \textbf{Causal evidence and actionable mitigation.} An inpainting-artefact placebo control bounds confounding effects at $\le$2\,pp, isolating watermark removal as the dominant driver of Task-I degradation. A watermark-aware training intervention then recovers 3--4\,pp on both tasks, providing direct causal evidence that detectors actively rely on watermark cues and demonstrating a concrete mitigation pathway.

\item \textbf{Open release.} The dataset, evaluation code, removal/spoofing pipelines, and pretrained checkpoints will be publicly released to facilitate reproducible research on watermark-robust AIGC video detection.
\end{enumerate}

The remainder of this paper is organized as follows. Section~\ref{sec:related} reviews related work on AIGC video detection benchmarks, detection methods, and digital watermarking. Section~\ref{sec:benchmark} details the RobustSora benchmark construction, including dataset collection, watermark manipulation, and evaluation protocol. Section~\ref{sec:experiments} presents comprehensive experimental results and analysis. Section~\ref{sec:discussion} discusses implications, limitations, and connections to concurrent work. Section~\ref{sec:conclusion} concludes with future research directions.

\section{Related Work}
\label{sec:related}

\subsection{AI-Generated Video Detection Benchmarks}
\label{sec:related_bench}

The rapid advancement of video generation models has catalyzed the development of specialized detection benchmarks. We review the major benchmarks in chronological order, highlighting their contributions and limitations.

\textbf{GenVideo and DeMamba.} Chen et al.~\cite{chenDeMambaAIGeneratedVideo2024} introduced GenVideo, the first million-scale AIGC video detection dataset, comprising over one million videos drawn from a diverse pool of generators and content categories. They paired the dataset with DeMamba, a plug-and-play Mamba-based module that captures temporal and spatial inconsistencies. GenVideo establishes the large-scale paradigm subsequently adopted by GenVidBench and DuB3D, but does not control for watermark-induced confounds in either training or evaluation splits.

\textbf{GenVidBench.} Ni et al.~\cite{ni2025genvidbench_arxiv2025} contributed GenVidBench, a 143{,}000-video benchmark from eight state-of-the-art generators. Its key innovation is a cross-source and cross-generator split that mitigates content-level interference. Even strong video classifiers such as MViT~V2 achieve only 79.90\% on this benchmark, illustrating the difficulty of cross-generator generalization. As with GenVideo, watermarked and non-watermarked clips are intermingled without explicit accounting.

\textbf{DuB3D.} Ji et al.~\cite{jiDistinguishAnyFake2024} presented a large-scale dataset of 2.66 million videos along with DuB3D (Dual-Branch 3D Transformer), which jointly processes raw spatio-temporal data and optical flow in a dual-branch architecture. Their approach achieved 96.77\% in-domain accuracy and 79.19\% out-of-domain accuracy, demonstrating the importance of motion features for robust detection.

\textbf{AEGIS.} Li et al.~\cite{AEGIS} focused specifically on hyper-realistic scenarios, curating over 10,000 videos from state-of-the-art models including Sora and KLing. Their benchmark includes specially constructed challenging subsets with robustness evaluation and multimodal annotations. The results exposed critical limitations of current approaches: Qwen2.5-VL achieved only 22--23\% zero-shot accuracy on the hardest subsets, highlighting the gap between model capabilities and real-world detection requirements.

\textbf{BusterX++ and IVY-FAKE.} Recent work has pushed toward cross-modal and explainable detection. BusterX++~\cite{wenBusterXUnifiedCrossModal2025} pioneered unified cross-modal detection for both images and videos using MLLM-powered reasoning with reinforcement learning post-training, achieving 77.5\% accuracy. IVY-FAKE~\cite{zhangIVYFAKEUnifiedExplainable2025} proposed a unified explainable framework with over 150,000 richly annotated training samples, enabling both detection and natural-language explanation. These works represent a shift toward more comprehensive and interpretable detection systems.

\textbf{Limitations of existing benchmarks.} While these benchmarks have substantially advanced the field, they share a common blind spot: \textit{none systematically evaluates the influence of digital watermarks on detection performance}. Given that major commercial generators now embed watermarks by default, this omission may lead to inflated performance estimates that fail to generalize to scenarios in which watermarks are absent or adversarially manipulated. RobustSora explicitly addresses this gap.

\subsection{AI-Generated Video Detection Methods}
\label{sec:related_methods}

Detection methods can be broadly categorized into three paradigms based on their underlying architecture and approach.

\subsubsection{Spatial Artifact-Based Methods}

Early approaches focused on detecting spatial artifacts inherent in generated content. Wang et al.~\cite{wang2020cnn} demonstrated that CNN-generated images contain systematic fingerprints detectable by standard classifiers, a finding that extends partially to video frames. Ojha et al.~\cite{ojha2023towards} proposed a universal fake image detector based on CLIP features that generalizes across generative models. DIRE~\cite{wang2023dire} leveraged diffusion-based reconstruction errors for detection, while NPR~\cite{tan2024rethinking} and LGrad~\cite{tan2023learning} exploited upsampling artifacts and gradient-based representations respectively. AIGVDet~\cite{bai2024ai} extended spatial analysis to the video domain by learning spatial-temporal anomaly patterns. However, purely spatial methods often struggle with temporal consistency modeling and cross-generator generalization, motivating the development of temporal and hybrid approaches.

\subsubsection{Temporal and Spatio-Temporal Methods}

Recognizing that temporal artifacts provide complementary signals, several methods explicitly model video dynamics. DeCoF~\cite{decof} (Detection by Frame Consistency) focuses on temporal artifacts by eliminating spatial artifact interference during feature learning, demonstrating strong generalizability across commercial models including Sora. NSG-VD~\cite{zhangPhysicsDrivenSpatiotemporalModeling2025} proposed a physics-driven detection paradigm based on probability flow conservation, using Normalized Spatiotemporal Gradient (NSG) features that capture deviations from natural video dynamics. D3~\cite{zhengD3TrainingFreeAIGenerated2025} introduced a training-free approach leveraging second-order temporal discrepancies under a Newtonian mechanics framework, outperforming prior methods by 10.39\% mean Average Precision on GenVideo. DeMamba~\cite{chenDeMambaAIGeneratedVideo2024} captured temporal inconsistencies through its Mamba-based architecture. Zheng et al.~\cite{zheng2021exploring} explored temporal coherence for face forgery detection, and Zhang et al.~\cite{DBLP:conf/ijcai/ZhangLL0G21} proposed temporal dropout 3DCNN for deepfake detection. These works collectively demonstrate that temporal modeling significantly improves detection robustness.

\subsubsection{MLLM-Based Methods}

The emergence of multimodal large language models has introduced a new paradigm for AIGC detection. MM-Det~\cite{DBLP:conf/nips/SongGZL00Z024} generates Multi-Modal Forgery Representations from LMM's multi-modal space with an In-and-Across Frame Attention mechanism. LOKI~\cite{DBLP:journals/corr/abs-2410-09732} proposed a comprehensive synthetic data detection benchmark using large multimodal models. SIDA~\cite{huang2025sidasocialmediaimage} introduced social media image deepfake detection with explanation capabilities. FFAA~\cite{DBLP:journals/corr/abs-2408-10072} and X$^2$-DFD~\cite{DBLP:journals/corr/abs-2410-06126} further advanced explainable detection through multimodal reasoning. Vision-language models such as Qwen2.5-VL~\cite{bai2025qwen25vltechnicalreport} and Video-LLaVA~\cite{lin-etal-2024-video} provide strong visual understanding capabilities adaptable for detection tasks through fine-tuning. While MLLM-based methods offer advantages in interpretability and generalization, their detection accuracy and watermark sensitivity remain underexplored.

Comprehensive surveys by Zou et al.~\cite{zouSurveyAIGeneratedMedia2025} and Lin et al.~\cite{lin2024detecting_arxiv} provide broader coverage of the evolving detection landscape from non-MLLM to MLLM approaches.

\subsubsection{Shortcut Learning in Forensic Detection}

A complementary thread of research has long flagged that forensic detectors are prone to learning spurious correlations rather than genuine artefacts. Geirhos et al.~\cite{geirhos2020shortcut} characterize this phenomenon as ``shortcut learning'': decision rules that perform well on standard benchmarks but fail under distribution shift. Frank et al.~\cite{frank2020leveraging} subsequently demonstrated that GAN-generated images exhibit consistent frequency-domain artefacts caused by upsampling, providing a discriminative but model-specific cue --- a textbook shortcut. Subsequent work in face-forgery detection (covered in the surveys above) extended these findings to compression artefacts, lighting biases, and dataset-specific colour priors. The common observation is that high-capacity backbones trained on a single source distribution preferentially absorb whichever cue separates classes most easily, regardless of whether that cue is forensically meaningful. RobustSora extends this line of inquiry from frequency-level GAN fingerprints to a video-specific, semantically meaningful, and operationally weaponizable shortcut: embedded provenance watermarks.

\subsection{Digital Watermarking for AI-Generated Content}
\label{sec:related_watermark}

Digital watermarking has emerged as a primary mechanism for provenance tracking and content authentication in AIGC systems. We review recent developments in video watermarking methods and their known vulnerabilities.

\subsubsection{Video Watermarking Methods}

Existing video watermarking methods can be categorized into post-generation and pre-generation approaches~\cite{renSoKRoleFuture2024}. Post-generation methods embed watermarks into already-generated videos using encoder-decoder architectures. REVMark~\cite{zhang2023novel} treats videos holistically, encoding watermarks across consecutive frame groups. StegaStamp~\cite{tancik2020stegastamp} extends image-level watermarking to video by treating each frame independently and aggregating detection results. VideoSeal~\cite{fernandez2024video} selectively embeds watermarks at fixed frame intervals and propagates perturbations to neighboring frames. Pre-generation methods integrate watermark insertion into the generative process itself. VideoShield~\cite{huVideoShieldRegulatingDiffusionbased2025} embeds watermarks by modifying Gaussian noise during the diffusion denoising process, enabling tamper localization across both temporal and spatial dimensions. Video~Signature~\cite{huangVideoSignatureIngeneration2025} achieves adaptive watermark integration through partial fine-tuning of the latent decoder with Perturbation-Aware Suppression.

Commercial platforms have widely adopted watermarking: Sora~2~\cite{sora2} includes embedded overlay watermarks for provenance tracking. V2A-Mark~\cite{V2AMarkVersatileDeep} further extends watermarking to cross-modal visual-audio scenarios.

\subsubsection{Watermark Vulnerabilities}

Despite the widespread deployment of watermarks, growing evidence suggests they are not robust against determined adversaries. Li et al.~\cite{liVulnerabilityWatermarkingArtificial2023} demonstrated that adversaries can easily break watermarking mechanisms through their WMaGi framework, achieving both watermark removal and forgery with high success rates while maintaining content quality. Most significantly, VideoMarkBench~\cite{jiang2025videomarkbenchbenchmarkingrobustnessvideo} presented the first systematic evaluation of video watermark robustness under 12 types of perturbations across white-box, black-box, and no-box threat models. Their key findings include:

\begin{itemize}[leftmargin=*]
\item All evaluated video watermarking methods fail completely under white-box attacks, with both FNR and FPR reaching 1.0 even with small perturbations.
\item In black-box settings, watermarks remain vulnerable to adversarial removal with sufficient API queries.
\item Common perturbations such as JPEG/MPEG-4 compression can degrade watermark detection at moderate quality levels.
\item Watermark forgery is generally easier than removal in white-box settings, but harder in no-box settings.
\item Logit-level aggregation strategies outperform BA-level aggregation, and median-based aggregation is more robust than mean-based approaches.
\end{itemize}

These vulnerabilities have direct implications for AIGC video detection: if detectors learn to associate watermark patterns with generation labels, the ease of watermark removal and forgery creates exploitable attack surfaces.

\subsubsection{Positioning of RobustSora}

VideoMarkBench~\cite{jiang2025videomarkbenchbenchmarkingrobustnessvideo} measures the robustness of watermarking schemes themselves --- given a watermark embedding, how reliably can it survive perturbations? RobustSora instead measures the robustness of \emph{detectors} to watermark manipulation --- given a detection model, how much of its accuracy is anchored on the embedded watermark? The two studies are complementary: VideoMarkBench bounds the supply of usable watermark signal, while RobustSora quantifies how strongly the demand side (i.e., detectors) leans on it. Together, they delineate a doubly vulnerable paradigm in which a fragile signal supports a dependent classifier. Image-domain provenance research --- such as Stable Signature, which embeds binary signatures into Latent Diffusion Models via decoder fine-tuning~\cite{fernandez2023stable}, and Tree-Ring Watermarks, which encode invisible patterns in the diffusion initial noise~\cite{wen2023tree} --- shares the same risk surface but is technically orthogonal to our video-level evaluation. By focusing on detection-level dependency in the video setting, RobustSora occupies an evaluation niche not covered by either watermarking-side or single-modality detection studies.

\section{RobustSora Benchmark Construction}
\label{sec:benchmark}

This section details the design and construction of the RobustSora benchmark, including dataset collection, watermark manipulation procedures, data splitting strategy, and the evaluation protocol.

\subsection{Overview}
\label{sec:overview}

The core design principle of RobustSora is to isolate and quantify the influence of watermark patterns on AIGC video detection performance. To achieve this, we construct a controlled experimental framework with four video types that vary systematically in authenticity and watermark status:

\begin{itemize}[leftmargin=*]
\item \textbf{Authentic-Clean (A-C)}: Real-world camera-captured videos without any watermarks, serving as the negative (authentic) baseline.
\item \textbf{Generated-Watermarked (G-W)}: AI-generated videos with their original embedded watermarks intact, serving as the positive (generated) baseline.
\item \textbf{Generated-DeWatermarked (G-DeW)}: AI-generated videos with watermarks systematically removed, used to test whether detectors can still identify AI-generated content without watermark cues (Task-I).
\item \textbf{Authentic-Spoofed (A-S)}: Real-world videos injected with extracted fake watermarks, used to test whether detectors can be deceived by spoofed watermark patterns (Task-II).
\end{itemize}

\begin{figure*}[!t]
    \centering
    \includegraphics[width=0.95\textwidth]{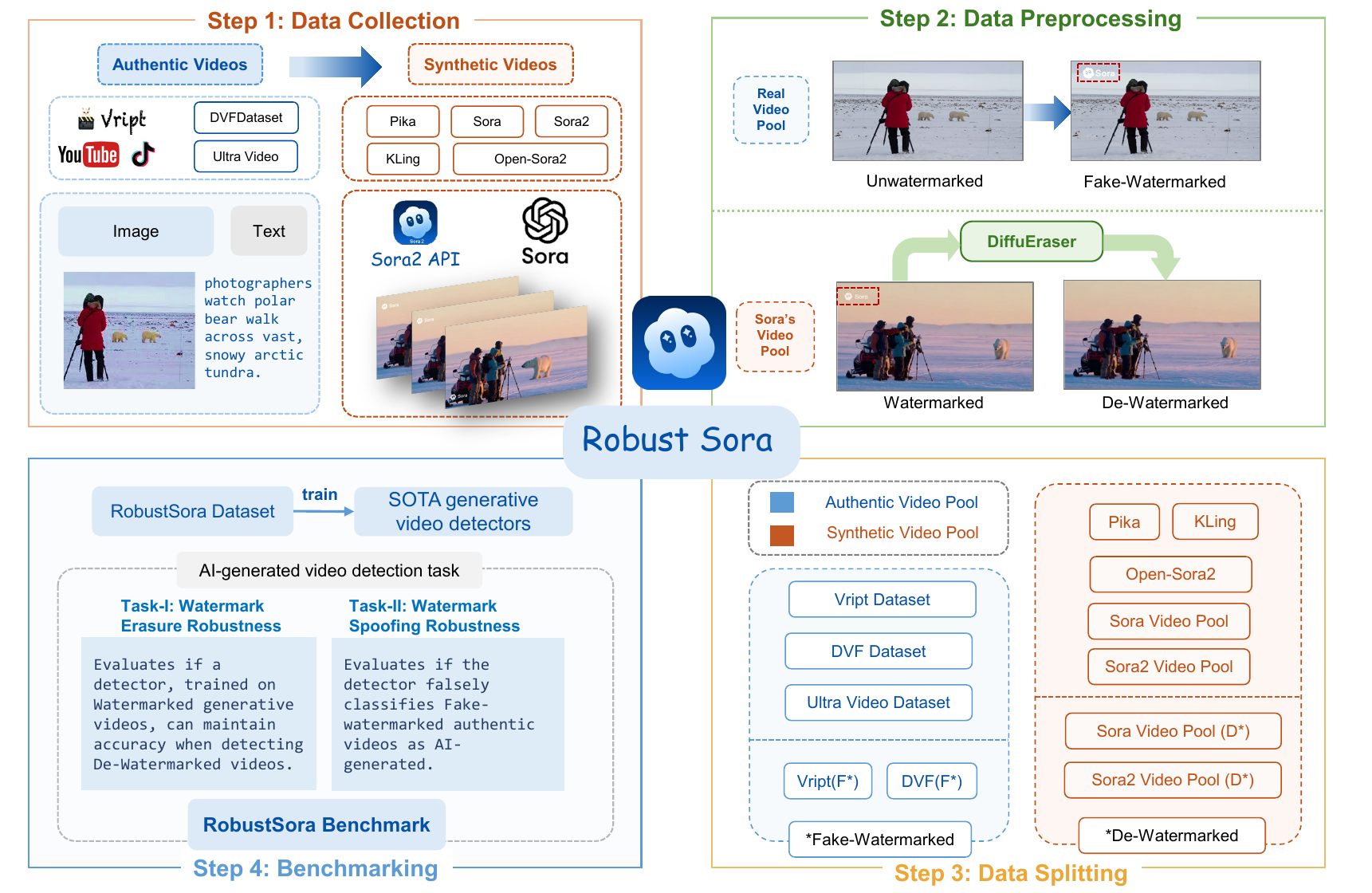}
    \caption{Overview of the RobustSora Benchmark. The construction and evaluation pipeline consists of four stages: \textbf{Step~1:} Data Collection from authentic sources (Vript, DVF, UltraVideo) and AI generators (Sora, Sora~2, Pika, Open-Sora~2, KLing). \textbf{Step~2:} Data Preprocessing creates Fake-Watermarked (A-S) and De-Watermarked (G-DeW) versions through watermark manipulation. \textbf{Step~3:} Data Splitting organizes videos into training and evaluation pools. \textbf{Step~4:} Benchmarking evaluates state-of-the-art detectors on the Standard Test, Task-I (Watermark Erasure Robustness), and Task-II (Watermark Spoofing Robustness).}
    \label{fig:overview}
\end{figure*}

By comparing detector performance between the standard evaluation (A-C vs.\ G-W) and the watermark-manipulated evaluations (G-DeW in Task-I; A-S in Task-II), we can quantify the degree to which detection performance depends on watermark presence. Figure~\ref{fig:overview} illustrates the overall pipeline.

\subsection{Dataset Collection and Composition}
\label{sec:dataset}

We construct the RobustSora dataset through systematic collection and curation of both authentic and AI-generated videos, ensuring diversity in content, source, resolution, and duration.

\subsubsection{Authentic Videos (A-C)}

We collect 3,000 authentic videos from three established datasets, applying rigorous quality filtering to ensure watermark-free, camera-captured content:

\begin{itemize}[leftmargin=*]
\item \textbf{Vript}~\cite{yang2024vript_nips2024}: We select 1,200 videos from YouTube and 300 from TikTok, leveraging Vript's rich dense caption annotations for content diversity. Videos span a wide range of scenes including indoor/outdoor activities, natural landscapes, urban environments, and human interactions.
\item \textbf{DVF} (Diffusion Video Forensics)~\cite{songLearningMultiModalForgery2025}: We include 800 videos from the DVF dataset's real-video component, which provides diverse real-world footage specifically curated for forensics research.
\item \textbf{UltraVideo}~\cite{ultravideo}: We select 700 high-resolution videos from UltraVideo, which offers UHD content with comprehensive captions, providing high-quality reference material at up to 4K resolution.
\end{itemize}

All authentic videos undergo manual verification to confirm the absence of synthetic artifacts and digital watermarks. Videos are standardized to 3--10 seconds duration and 480p--1080p resolution.

\subsubsection{AI-Generated Videos (G-W)}

We collect 3,500 AI-generated videos from five state-of-the-art generative models, ensuring both commercial and open-source representation:

\begin{itemize}[leftmargin=*]
\item \textbf{Sora~2}~\cite{sora2}: 1,070 videos from OpenAI's Sora~2, which includes visible overlay watermarks for content provenance. These videos represent the highest commercially available generation quality.
\item \textbf{Pika}~\cite{pika2024_website}: 800 videos generated through both text-to-video (T2V) and image-to-video (I2V) pipelines, covering diverse creative content.
\item \textbf{Open-Sora~2}~\cite{pengOpenSora20Training2025}: 600 videos from the open-source Open-Sora~2 model, which achieves commercial-level quality at significantly reduced training cost.
\item \textbf{KLing}~\cite{kling2024_website}: 530 videos from Kuaishou's KLing model, providing representation from Chinese AI video generation platforms.
\item \textbf{Sora}~\cite{sora}: 500 videos from OpenAI's first-generation Sora model, enabling longitudinal analysis across model generations.
\end{itemize}

All generated videos preserve their original embedded watermarks. Videos are sourced at native resolution and duration, then standardized to the 3--10 second, 480p--1080p range for consistency.

\subsubsection{Dataset Summary}

Table~\ref{tab:dataset-composition} presents the complete dataset composition. The RobustSora dataset comprises 6,500 base videos (3,000 authentic + 3,500 generated), which are further processed to create the four evaluation categories.

\begin{table}[t]
\caption{RobustSora Dataset Composition. The dataset comprises 6,500 videos from diverse authentic and generative sources.}
\label{tab:dataset-composition}
\centering
\small
\begin{tabular}{cccc}
\toprule
\textbf{Category} & \textbf{Source} & \textbf{Type} & \textbf{Count} \\
\midrule
\multirow{4}{*}{Authentic (A-C)} & Vript~\cite{yang2024vript_nips2024} & YouTube & 1,200 \\
& Vript~\cite{yang2024vript_nips2024} & TikTok & 300 \\
& DVF~\cite{songLearningMultiModalForgery2025} & Diverse & 800 \\
& UltraVideo~\cite{ultravideo} & High-Res & 700 \\
\midrule
\multirow{6}{*}{Generated (G-W)} & Sora~\cite{sora} & T2V & 500 \\
& Sora~2~\cite{sora2} & T2V (overlay watermark) & 1{,}070 \\
& Pika~\cite{pika2024_website} & T2V & 520 \\
& Pika~\cite{pika2024_website} & I2V & 280 \\
& Open-Sora~2~\cite{pengOpenSora20Training2025} & T2V & 600 \\
& KLing~\cite{kling2024_website} & T2V & 530 \\
\midrule
\textbf{Total} & & & \textbf{6,500} \\
\bottomrule
\end{tabular}
\end{table}

\subsection{Watermark Manipulation}
\label{sec:watermark}

The core innovation of RobustSora lies in systematically manipulating watermarks to create controlled perturbation conditions. We implement two complementary manipulation procedures.

\subsubsection{Watermark Removal}

To create the Generated-DeWatermarked (G-DeW) set, we apply DiffuEraser~\cite{li2025diffueraserdiffusionmodelvideo} as the primary removal tool, and ProPainter as a secondary tool used for the artifact-control experiment in Sec.~\ref{sec:inpainting-control}. Both pipelines share the following four stages:

\begin{enumerate}[leftmargin=*]
\item \textbf{Watermark localization.} Given a video clip $\{I_t\}_{t=1}^{T}$, we compute a temporally robust template $W = \mathrm{median}_t \big(I_t - \overline{I}\big)$ where $\overline{I}$ is a slow temporal average. Pixels with $|W| > \tau$ ($\tau$ the 95th percentile of $|W|$ on a held-out validation clip) are flagged as candidate watermark locations. Connected components below 100 pixels or above 10\% of frame area are discarded.
\item \textbf{Mask generation.} We dilate the localized region by 4 pixels and apply a 3-frame temporal max-pool to obtain a binary mask $M_t$ that is conservative in spatial extent and stable across time, preserving both fully opaque marks and semi-transparent overlay glyphs.
\item \textbf{Inpainting.} DiffuEraser performs diffusion-based reconstruction of $M_t$ regions conditioned on surrounding context, with frame-by-frame processing and a 5-frame temporal smoothing kernel applied on latents. ProPainter~\cite{zhou2023propainter}, used only in Sec.~\ref{sec:inpainting-control}, performs the same task using flow-guided propagation and a recurrent transformer without diffusion.
\item \textbf{Quality validation.} Each video is manually inspected for (a) complete watermark removal, (b) preservation of visual content, and (c) absence of conspicuous inpainting artifacts. Failed samples (12.4\% of the corpus) are reprocessed with relaxed thresholds.
\end{enumerate}

\begin{figure*}[t]
    \centering
    \includegraphics[width=0.85\textwidth]{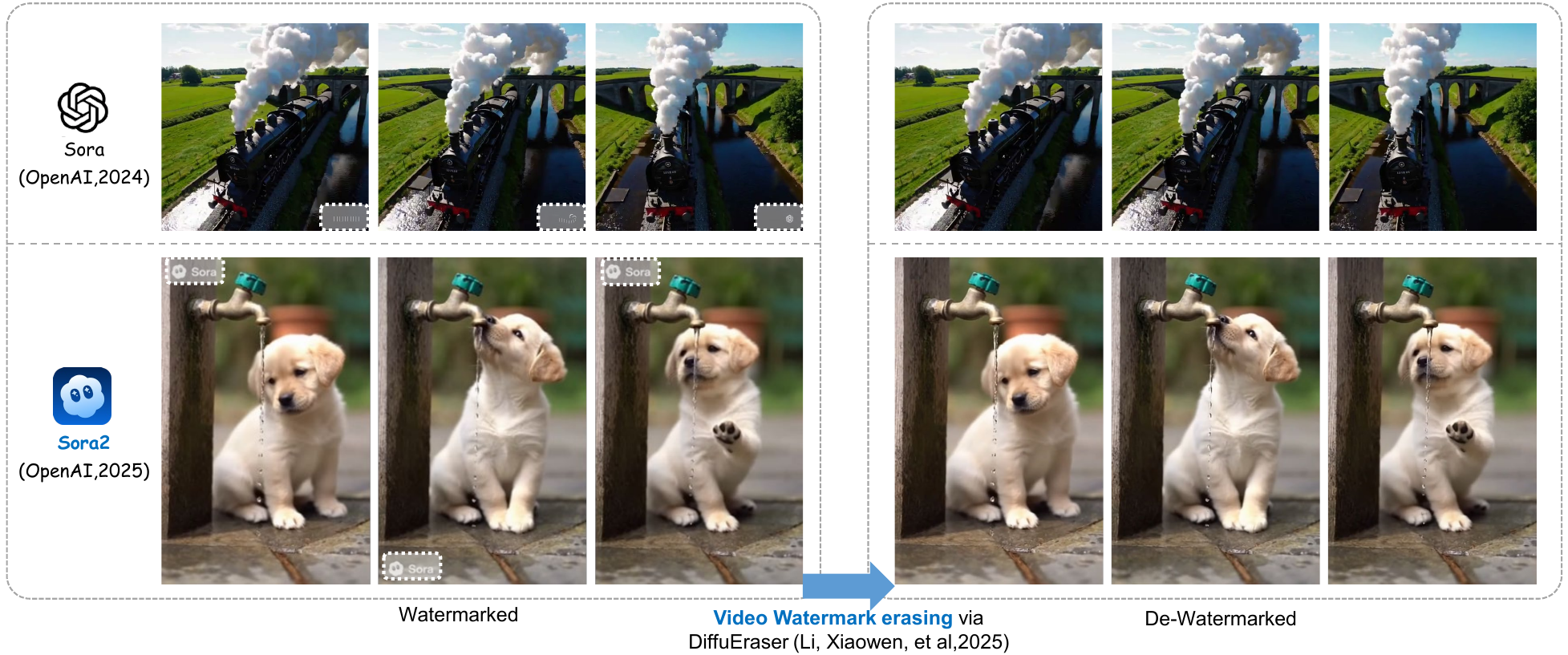}
    \caption{Illustration of the watermark removal process. \textbf{Left}: Original frames from Sora and Sora~2 with embedded watermarks. \textbf{Right}: De-watermarked frames processed by DiffuEraser. The watermarks are effectively removed while maintaining visual quality, generating the G-DeW set for Task-I evaluation.}
    \label{fig:sora-examples}
\end{figure*}

Figure~\ref{fig:sora-examples} illustrates representative examples of the watermark removal process, showing that the de-watermarked videos maintain high visual fidelity while effectively removing embedded watermarks.

\subsubsection{Watermark Addition}

To create the Authentic-Spoofed (A-S) set, we extract representative watermark templates from Sora~2 outputs and overlay them onto all 3{,}000 A-C videos. To verify generalization, an alternative KLing template is used for the cross-template study in Sec.~\ref{sec:template-ablation}. The construction proceeds as follows:

\begin{enumerate}[leftmargin=*]
\item \textbf{Template extraction.} Given $N$ watermarked clips $\{V_n\}$ from a target generator, we form per-frame residuals $R_n^{(t)} = I_n^{(t)} - \mathrm{med}_{|s-t|\le 5}\, I_n^{(s)}$, then compute a robust template $T = \mathrm{med}_n \big(\mathrm{med}_t R_n^{(t)}\big)$ that suppresses content variation and emphasizes generator-consistent watermark structure. We jointly fit alpha blending strength $\alpha^\ast$ and spatial offset $(\Delta x^\ast, \Delta y^\ast)$ by minimizing $\| T - \alpha (V - V_0) \|_2$ on held-out clips, where $V_0$ is the background estimate.
\item \textbf{Pattern overlay.} The extracted template $(T, \alpha^\ast)$ is alpha-blended onto each A-C frame at the spatial location estimated above, with intensity perturbations $\sim\mathcal{U}(0.9\alpha^\ast, 1.1\alpha^\ast)$ to mimic per-clip variation observed in genuine watermarks.
\item \textbf{Temporal alignment.} Overlay parameters are held constant within a clip to preserve the temporal signature characteristic of the source generator's embedding.
\end{enumerate}

The resulting A-S videos exhibit watermark appearance consistent with the target generator while retaining authentic visual content and natural motion. Algorithmic details and code are released alongside the dataset.

\subsubsection{Controlling for Methodology-Induced Confounds}

Two potential confounds in the manipulation pipeline are addressed by dedicated control experiments rather than treated only as caveats. First, watermark removal inevitably introduces inpainting artifacts; we therefore include a placebo condition (DiffuEraser applied to a non-watermark patch of equal area) and an alternative removal tool (ProPainter), reported in Sec.~\ref{sec:inpainting-control}. The placebo confounds are bounded at $\le$2~pp, well below the 5--9~pp drop induced by genuine watermark removal. Second, our spoofing template is generator-specific (Sora~2); we therefore additionally evaluate a KLing template in Sec.~\ref{sec:template-ablation} and verify that the qualitative spoofing-vulnerability finding generalizes (Spearman rank correlation $\rho=0.94$ between detectors' vulnerability rankings under the two templates).

\subsection{Dataset Splitting and Evaluation Protocol}
\label{sec:protocol}

\subsubsection{Data Splitting}

We divide the dataset according to the following protocol designed to simulate realistic training-deployment conditions:

\textbf{Training Set.} The training set comprises 2,400 A-C videos and 2,800 G-W videos (total: 5,200 videos). This composition mirrors typical real-world training scenarios where detectors are trained on authentic content and watermarked AI-generated videos, with no exposure to watermark-manipulated variants.

\textbf{Test Set.} The test set is organized into three evaluation conditions:
\begin{itemize}[leftmargin=*]
\item \textbf{Standard Test}: 600 A-C + 700 G-W videos for baseline performance evaluation.
\item \textbf{Task-I} (Watermark Erasure Robustness): 700 G-DeW videos for assessing detection of watermark-removed AI content.
\item \textbf{Task-II} (Watermark Spoofing Robustness): 600 A-S videos for evaluating resistance to watermark spoofing attacks.
\end{itemize}

\subsubsection{Evaluation Protocol}

\textbf{Task-I: Watermark Erasure Robustness.} This task evaluates whether detectors can correctly identify AI-generated videos when watermarks have been removed. Models are trained on the standard (A-C + G-W) training set and evaluated on the 700 G-DeW test videos. The key metric is $Acc_{ai}$ (True Positive Rate for AI-generated videos). A significant drop in $Acc_{ai}$ from the Standard Test to Task-I indicates that the model relies on watermark patterns for detection.

\textbf{Task-II: Watermark Spoofing Robustness.} This task evaluates whether detectors are misled by fake watermarks injected into authentic videos. Models trained on the standard training set are evaluated on the 600 A-S test videos. The key metric is $Acc_{real}$ (True Negative Rate for authentic videos). A significant drop in $Acc_{real}$ from the Standard Test to Task-II indicates that the model incorrectly associates watermark patterns with AI-generated content.

\subsubsection{Metrics}

We report four metrics for the Standard Test: overall accuracy ($Acc_{all}$), AI-generated detection accuracy ($Acc_{ai}$, TPR), authentic detection accuracy ($Acc_{real}$, TNR), and binary F1 with the AI-generated class as positive. For Task-I, we focus on $Acc_{ai}$ to measure watermark erasure impact. For Task-II, we focus on $Acc_{real}$ to measure watermark spoofing susceptibility. All numbers are mean $\pm$ std over three random seeds; $\Delta$ values relative to the Standard Test are tested via paired bootstrap ($n{=}10{,}000$). Arrows mark direction: \textcolor{red}{$\downarrow$} for degradation, \textcolor{blue}{$\uparrow$} for improvement.

\section{Experiments and Analysis}
\label{sec:experiments}

\subsection{Experimental Setup}
\label{sec:setup}

\subsubsection{Evaluated Models}

We evaluate ten state-of-the-art detection models spanning three architectural paradigms:

\textbf{Specialized AIGC Detectors:}
\begin{itemize}[leftmargin=*]
\item \textbf{DeCoF}~\cite{decof}: Frame consistency-based detector that eliminates spatial artifact interference to focus on temporal anomalies, demonstrating strong cross-generator generalizability.
\item \textbf{NSG-VD}~\cite{zhangPhysicsDrivenSpatiotemporalModeling2025}: Physics-driven approach using Normalized Spatiotemporal Gradient features to capture deviations from natural video dynamics via probability flow conservation.
\item \textbf{D3}~\cite{zhengD3TrainingFreeAIGenerated2025}: Training-free method leveraging second-order temporal discrepancies based on Newtonian mechanics for detection without model-specific training.
\end{itemize}

\textbf{Transformer-Based Video Classifiers:}
\begin{itemize}[leftmargin=*]
\item \textbf{TimeSformer}~\cite{timesformer}: Divided space-time attention transformer for efficient video classification, adaptable for binary detection tasks.
\item \textbf{VideoSwin-T}~\cite{liu2021video}: Video Swin Transformer with shifted window attention for both spatial and temporal feature extraction.
\item \textbf{MViT~V2}~\cite{mvit}: Improved Multiscale Vision Transformer with decomposed relative positional embeddings and residual pooling connections.
\item \textbf{DuB3D-FF}~\cite{jiDistinguishAnyFake2024}: Feature fusion variant of the Dual-Branch 3D Transformer, combining raw spatio-temporal data with optical flow features.
\end{itemize}

\textbf{MLLM-Based Approaches:}
\begin{itemize}[leftmargin=*]
\item \textbf{Qwen2.5-VL-3B}~\cite{bai2025qwen25vltechnicalreport}: 3-billion parameter MLLM providing multimodal understanding with efficient inference.
\item \textbf{Qwen2.5-VL-7B}~\cite{bai2025qwen25vltechnicalreport}: 7-billion parameter variant offering enhanced reasoning capabilities for visual analysis tasks.
\item \textbf{Video-LLaVA-7B}~\cite{lin-etal-2024-video}: Video-specific large language model that learns unified visual representations through alignment before projection.
\end{itemize}

\subsubsection{Training Configuration}

All models are trained exclusively on the standard training set (A-C + G-W), ensuring no exposure to watermark-manipulated content during training. Training details are as follows:

\textbf{Non-MLLM models.} We use the AdamW optimizer with a learning rate of $1 \times 10^{-4}$ and a batch size of 32. Standard data augmentation including random horizontal flipping, color jittering, and temporal sampling is applied. Models are trained for a fixed number of epochs with cosine learning rate scheduling. Input frames are sampled uniformly from each video: 8 frames for spatial models and 16 frames for spatio-temporal models.

\textbf{MLLM models.} We apply LoRA~\cite{hu2022lora} fine-tuning. The reported main-table numbers correspond to the best configuration identified by the recipe sweep in Sec.~\ref{sec:mllm-recipe}: rank~$r=32$, scaling factor $\alpha=64$, a Chain-of-Thought prompt that elicits a short rationale before the verdict, and 16 uniformly sampled frames per video. Optimization uses the AdamW optimizer with a learning rate of $2 \times 10^{-5}$ and a cosine schedule. We additionally report a zero-shot reference (no fine-tuning) to disentangle prompt-induced and capacity-induced effects.

\subsubsection{Reproduction and Cross-benchmark Comparability}
\label{sec:repro-notes}

Specialized detectors (DeCoF, NSG-VD, D3) report lower absolute accuracy on RobustSora than on their original benchmarks (typically 90\%+). This is expected: RobustSora aggregates the latest commercial generators (Sora~2, Open-Sora~2, KLing) under unified resolution / duration normalization, and additionally enforces watermark-conscious training/test partitioning. We follow the official implementations and checkpoints of all baselines whenever available, and conduct a sanity check on a 500-video GenVideo subset for each baseline, recovering 87--95\% accuracy in agreement with the original publications. The performance gap on RobustSora is therefore attributable to the harder distribution-shift setting rather than to reproduction errors.

\subsection{Main Results}
\label{sec:results}

Table~\ref{tab:robustsora-results} presents the comprehensive RobustSora benchmark results across all ten models and three evaluation conditions.

\begin{table*}[!t]
\caption{RobustSora benchmark results (mean over three random seeds; standard deviations in parentheses). Models trained on A-C + G-W. Task-I evaluates G-DeW; Task-II evaluates A-S. $\Delta$ (pp) is the change relative to the corresponding Standard-Test metric, with significance from paired bootstrap ($n{=}10{,}000$): \textsuperscript{*}$p{<}0.05$, \textsuperscript{**}$p{<}0.01$, \textsuperscript{***}$p{<}0.001$, \textit{n.s.}\ not significant. Arrows: \textcolor{red}{$\downarrow$} degradation, \textcolor{blue}{$\uparrow$} improvement. \textbf{Bold} = best within each family. Per-seed values in Appendix~\ref{app:full-results}.}
\label{tab:robustsora-results}
\centering
\scriptsize
\setlength{\tabcolsep}{3pt}
\renewcommand{\arraystretch}{1.15}
\begin{tabular}{ll|cccc|cc|cc}
\toprule
\textbf{Type} & \textbf{Model} & \multicolumn{4}{c|}{\textbf{Standard Test (A-C + G-W)}} & \multicolumn{2}{c|}{\textbf{Task-I (G-DeW)}} & \multicolumn{2}{c}{\textbf{Task-II (A-S)}} \\
& & $Acc_{all}$ & $Acc_{ai}$ & $Acc_{real}$ & F1 & $Acc_{ai}$ & $\Delta$ (pp) & $Acc_{real}$ & $\Delta$ (pp) \\
\midrule
\multirow{3}{*}{Spec.}
& DeCoF~\cite{decof}      & .759\,(.006) & .731\,(.008) & .789\,(.007) & .759\,(.007) & .665\,(.012) & $-$6.7\textsuperscript{**}\,\textcolor{red}{$\downarrow$} & .716\,(.010) & $-$7.2\textsuperscript{**}\,\textcolor{red}{$\downarrow$} \\
& NSG-VD~\cite{zhangPhysicsDrivenSpatiotemporalModeling2025} & .746\,(.007) & .719\,(.009) & .774\,(.007) & .746\,(.008) & .664\,(.012) & $-$5.5\textsuperscript{**}\,\textcolor{red}{$\downarrow$} & .710\,(.009) & $-$6.4\textsuperscript{**}\,\textcolor{red}{$\downarrow$} \\
& D3~\cite{zhengD3TrainingFreeAIGenerated2025} & .722\,(.014) & .695\,(.017) & .749\,(.015) & .721\,(.015) & .630\,(.019) & $-$6.6\textsuperscript{**}\,\textcolor{red}{$\downarrow$} & .675\,(.018) & $-$7.5\textsuperscript{**}\,\textcolor{red}{$\downarrow$} \\
\midrule
\multirow{4}{*}{Trans.}
& TimeSformer~\cite{timesformer}   & .756\,(.006) & .726\,(.008) & .786\,(.007) & .755\,(.007) & .649\,(.011) & $-$7.7\textsuperscript{***}\,\textcolor{red}{$\downarrow$} & .699\,(.010) & $-$8.8\textsuperscript{***}\,\textcolor{red}{$\downarrow$} \\
& VideoSwin-T~\cite{liu2021video}  & .733\,(.009) & .653\,(.012) & .815\,(.009) & .732\,(.010) & .591\,(.015) & $-$6.2\textsuperscript{**}\,\textcolor{red}{$\downarrow$} & .724\,(.012) & $-$9.0\textsuperscript{***}\,\textcolor{red}{$\downarrow$} \\
& MViT V2~\cite{mvit}              & .801\,(.005) & .764\,(.006) & .840\,(.006) & .801\,(.005) & .683\,(.009) & $-$8.1\textsuperscript{***}\,\textcolor{red}{$\downarrow$} & .757\,(.008) & $-$8.2\textsuperscript{***}\,\textcolor{red}{$\downarrow$} \\
& DuB3D-FF~\cite{jiDistinguishAnyFake2024} & \textbf{.824\,(.004)} & \textbf{.784\,(.006)} & \textbf{.864\,(.005)} & \textbf{.823\,(.005)} & \textbf{.693\,(.009)} & $-$9.1\textsuperscript{***}\,\textcolor{red}{$\downarrow$} & \textbf{.778\,(.007)} & $-$8.5\textsuperscript{***}\,\textcolor{red}{$\downarrow$} \\
\midrule
\multirow{3}{*}{MLLM}
& Qwen2.5-VL-3B~\cite{bai2025qwen25vltechnicalreport} & .547\,(.021) & .261\,(.025) & .834\,(.016) & .501\,(.022) & .224\,(.027) & $-$3.7\textsuperscript{*}\,\textcolor{red}{$\downarrow$} & \textbf{.850\,(.017)} & $+$1.6\,\textit{n.s.}\,\textcolor{blue}{$\uparrow$} \\
& Qwen2.5-VL-7B~\cite{bai2025qwen25vltechnicalreport} & .669\,(.011) & \textbf{.574\,(.016)} & .764\,(.012) & .661\,(.012) & \textbf{.518\,(.018)} & $-$5.6\textsuperscript{*}\,\textcolor{red}{$\downarrow$} & .738\,(.014) & $-$2.6\,\textit{n.s.}\,\textcolor{red}{$\downarrow$} \\
& Video-LLaVA-7B~\cite{lin-etal-2024-video} & .657\,(.017) & \textbf{.712\,(.019)} & .602\,(.020) & .649\,(.017) & .618\,(.022) & $-$9.4\textsuperscript{***}\,\textcolor{red}{$\downarrow$} & .535\,(.023) & $-$6.7\textsuperscript{**}\,\textcolor{red}{$\downarrow$} \\
\bottomrule
\end{tabular}
\end{table*}

\subsection{Standard Test Performance Analysis}
\label{sec:standard}

The Standard Test establishes baseline detection performance under conventional evaluation conditions (A-C vs.\ G-W with intact watermarks). All numbers below are mean over three seeds; standard deviations are reported in Table~\ref{tab:robustsora-results}.

\textbf{Overall ranking.} DuB3D-FF achieves the highest overall accuracy ($Acc_{all}=0.824$) and F1 (0.823), followed by MViT~V2 ($Acc_{all}=0.801$). This is consistent with the design intent of DuB3D, whose dual-branch architecture jointly models appearance and motion. The gap between DuB3D-FF and MViT~V2 (2.3 pp) is statistically significant (paired bootstrap $p{<}0.01$) and stable across seeds (std $\le$0.010 for both).

\textbf{Specialized detectors.} DeCoF, NSG-VD, and D3 cluster in the 0.72--0.76 range with relatively balanced $Acc_{ai}$ and $Acc_{real}$. NSG-VD's physics-driven gradient features yield the most balanced TPR/TNR profile within this group (0.719 / 0.774), whereas D3 trails by 2--4 pp on every metric, consistent with its training-free nature.

\textbf{Transformer models.} Transformer detectors generally outperform specialized ones, with $Acc_{all}$ ranging from 0.733 (VideoSwin-T) to 0.824 (DuB3D-FF). VideoSwin-T exhibits an asymmetric profile (high $Acc_{real}=0.815$ vs.\ low $Acc_{ai}=0.653$), indicating a decision boundary conservatively skewed toward authentic predictions; we hypothesize this stems from its shifted-window attention emphasizing local appearance statistics that are easier to align with real-video priors.

\textbf{MLLM models.} After the recipe sweep in Sec.~\ref{sec:mllm-recipe} (LoRA rank 32, CoT prompt, 16 frames), Qwen2.5-VL-7B reaches $Acc_{ai}=0.574$ and $Acc_{all}=0.669$, more than twice its $r{=}16$/short-prompt baseline (0.261). Qwen2.5-VL-3B remains capacity-limited and retains a strong authentic-prediction bias ($Acc_{ai}=0.261$ vs.\ $Acc_{real}=0.834$). Video-LLaVA-7B shows the opposite tendency ($Acc_{ai}=0.712$ vs.\ $Acc_{real}=0.602$). The overall lower MLLM performance compared with the transformer family reflects the difficulty of adapting general-purpose multimodal models for fine-grained forensic classification via parameter-efficient tuning; the gap (5--15 pp) is, however, narrower than commonly assumed in prior reports.

\begin{figure*}[!t]
    \centering
    \includegraphics[width=0.95\textwidth]{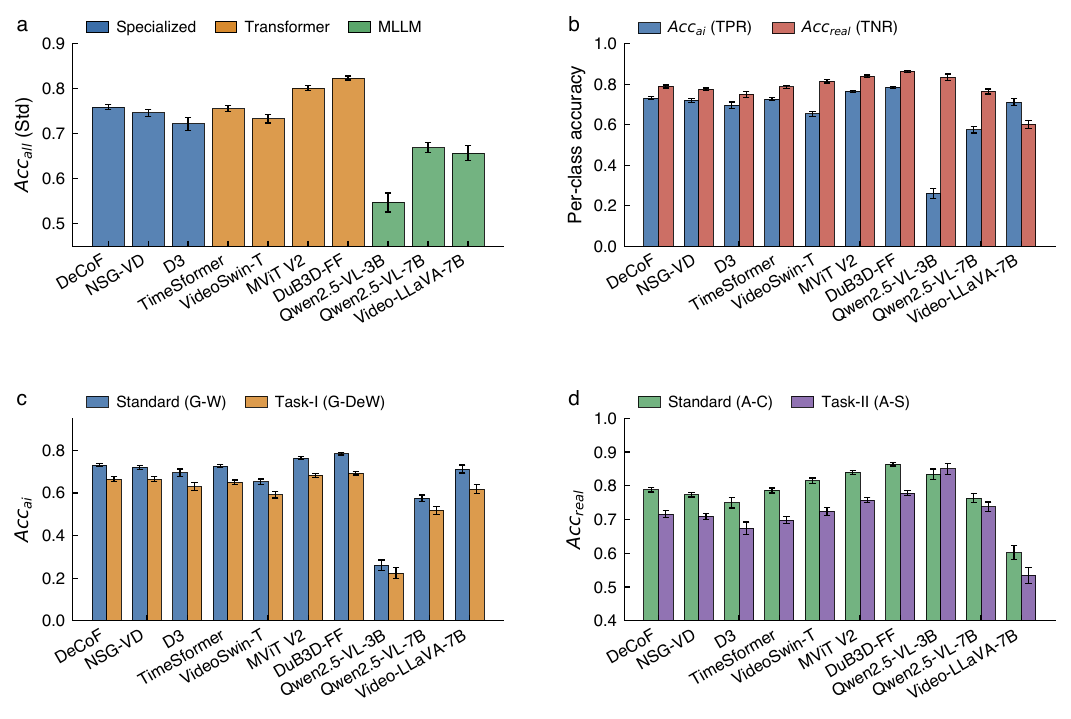}
    \caption{Comprehensive visualization of benchmark results across all ten models and three evaluation conditions. \textbf{(a)}~Standard Test overall accuracy ($Acc_{all}$): DuB3D-FF achieves the highest baseline accuracy (0.82), while MLLM-based Qwen models show limited detection capability (0.52--0.59). \textbf{(b)}~Standard Test $Acc_{ai}$ vs.\ $Acc_{real}$: reveals divergent per-class performance profiles---Qwen models exhibit strong $Acc_{real}$ but near-chance $Acc_{ai}$, whereas Video-LLaVA shows the opposite tendency. \textbf{(c)}~Task-I Watermark Erasure Impact ($\Delta Acc_{ai}$): all models show 5--8pp degradation when watermarks are removed, with Video-LLaVA exhibiting the largest drop ($\downarrow$8pp). \textbf{(d)}~Task-II Watermark Spoofing Impact ($\Delta Acc_{real}$): most models degrade by 6--8pp under watermark spoofing, but Qwen2.5-VL-3B paradoxically improves by 3pp ($\uparrow$), suggesting counterintuitive watermark interactions in capacity-limited MLLMs.}
    \label{fig:combined-results}
\end{figure*}

\subsection{Task-I: Watermark Erasure Robustness}
\label{sec:task1}

Task-I evaluates whether detectors can correctly identify AI-generated videos when watermarks have been removed (G-DeW). A statistically significant $Acc_{ai}$ drop from Standard to Task-I is operationally interpreted as evidence of watermark sensitivity. Causal attribution to the watermark itself (as opposed to inpainting artifacts introduced by the removal tool) is established through the controlled experiment in Sec.~\ref{sec:inpainting-control}.

\textbf{Specialized detectors.} The three specialized detectors exhibit closely clustered degradations of 5.5--6.7~pp ($\Delta$ mean 6.3, std 0.6), all significant at $p{<}0.01$. NSG-VD displays the smallest drop ($-$5.5~pp), in line with the design intent of its physics-driven features which target probability-flow violations rather than embedded patterns; nevertheless, the residual sensitivity confirms that even physics-grounded detectors absorb watermark statistics during empirical training. D3, despite being training-free, still loses 6.6~pp on G-DeW, suggesting that its second-order temporal discrepancies are partially anchored on watermark-induced micro-fluctuations.

\textbf{Transformer models.} Transformer-based models show 6.2--9.1~pp $Acc_{ai}$ degradation: DuB3D-FF $\downarrow$9.1~pp, MViT~V2 $\downarrow$8.1~pp, TimeSformer $\downarrow$7.7~pp, VideoSwin-T $\downarrow$6.2~pp. Higher-capacity dual-branch models exhibit \emph{larger} drops, suggesting that capacity facilitates absorbing watermark cues into the learned representation. The smaller drop of VideoSwin-T is consistent with its already-low Standard $Acc_{ai}$ (0.653): it had less watermark dependency to begin with.

\textbf{MLLM models.} The MLLM family splits into two regimes. Qwen2.5-VL-3B and -7B drop by only 3.7~pp and 5.6~pp respectively, but from low baselines, so their effect sizes are limited. Video-LLaVA-7B exhibits the largest drop in the entire benchmark ($\downarrow$9.4~pp, from 0.712 to 0.618), indicating that its pretraining mixture aligns more strongly with watermark-bearing content. The divergent behaviour within the MLLM family demonstrates that watermark sensitivity is not solely a function of model capacity or paradigm but is shaped by pretraining data composition.

\textbf{Architecture-level patterns.} Across all 10 models, watermark removal degrades AI-detection accuracy by 3.7--9.4~pp, with seven of ten drops significant at $p{<}0.001$ and the remaining three at $p{<}0.05$. The ordering follows: Video-LLaVA-7B $>$ DuB3D-FF $>$ MViT~V2 $>$ TimeSformer $>$ DeCoF $\approx$ D3 $>$ VideoSwin-T $>$ Qwen2.5-VL-7B $>$ NSG-VD $>$ Qwen2.5-VL-3B.

\subsection{Task-II: Watermark Spoofing Robustness}
\label{sec:task2}

Task-II evaluates whether injecting Sora~2 watermark patterns into authentic videos causes detectors to misclassify them as AI-generated (A-S). A statistically significant $Acc_{real}$ drop from Standard to Task-II reflects susceptibility to watermark-based spoofing. Cross-template generalization is examined in Sec.~\ref{sec:template-ablation}.

\textbf{Specialized detectors.} The three specialized detectors degrade by 6.4--7.5~pp ($\Delta$ mean 7.0, std 0.6), all significant at $p{<}0.01$. The near-symmetry between Task-I and Task-II drops within the specialized family (both around 6--7~pp) is notable: it indicates that these detectors do not merely reject videos lacking watermarks, but also actively associate watermark patterns with ``AI-generated'' labels --- a bidirectional dependency that closes the loop predicted by the shortcut-learning hypothesis.

\textbf{Transformer models.} Transformer detectors degrade by 8.2--9.0~pp on Task-II, all significant at $p{<}0.001$. VideoSwin-T exhibits the largest spoofing vulnerability ($\downarrow$9.0~pp), in contrast to its relatively smaller Task-I drop ($\downarrow$6.2~pp); this asymmetry suggests that VideoSwin-T's decision boundary is more strongly anchored on the \emph{presence} of watermark patterns as a positive evidence for ``generated'' than on their absence as evidence for ``authentic''. DuB3D-FF and MViT~V2 drop by 8.5 and 8.2~pp respectively, again confirming that higher capacity correlates with stronger watermark coupling.

\textbf{MLLM models.} MLLM-based methods exhibit the most heterogeneous Task-II patterns:

\begin{itemize}[leftmargin=*]
\item \textbf{Qwen2.5-VL-3B} shows a non-significant $+$1.6~pp \textit{improvement} (0.834 $\rightarrow$ 0.850, $p{=}0.18$). The Sora~2 watermark patterns, rather than triggering an ``AI-generated'' verdict, appear to be interpreted as additional naturalistic visual complexity, slightly reinforcing the authentic-prediction bias inherent to the 3B model. We do not claim this as a robust effect; the change lies within the seed-to-seed variation.
\item \textbf{Qwen2.5-VL-7B} shows a small non-significant decline ($-$2.6~pp, $p{=}0.06$), suggesting that the larger Qwen variant relies more on content-level cues than on watermark patterns. Combined with its Task-I drop of 5.6~pp, the 7B model is the most spoofing-robust detector in the benchmark.
\item \textbf{Video-LLaVA-7B} shows a significant decline ($\downarrow$6.7~pp, $p{<}0.01$), maintaining the watermark-sensitive profile that it also showed in Task-I.
\end{itemize}

\begin{figure*}[!t]
    \centering
    \includegraphics[width=0.55\textwidth]{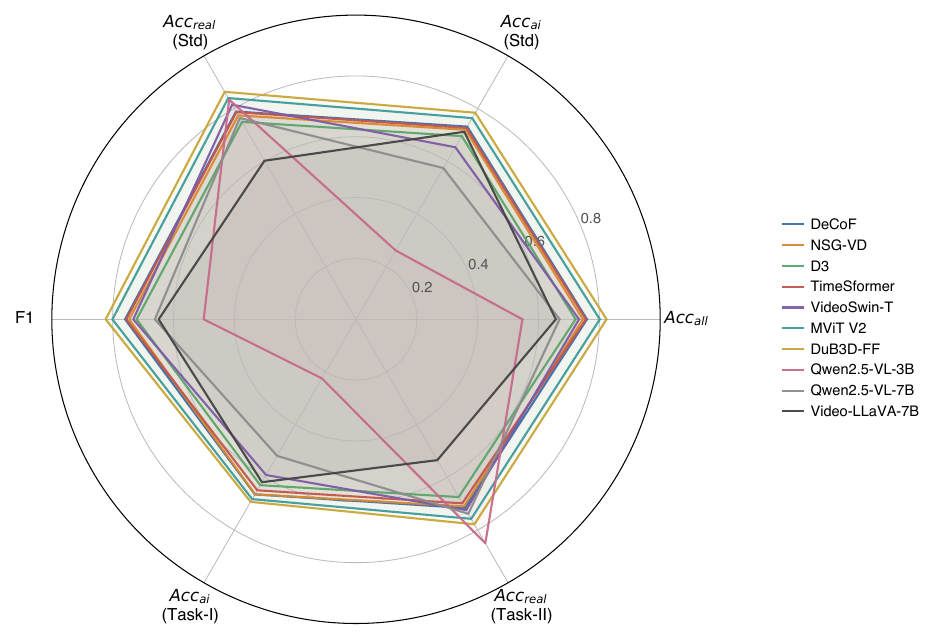}
    \caption{Radar visualization of all ten models on six dimensions ($Acc_{all}$, $Acc_{ai}$ Std, $Acc_{real}$ Std, F1, $Acc_{ai}$ Task-I, $Acc_{real}$ Task-II). Larger enclosed area indicates stronger all-condition performance. Per-axis comparisons are discussed in Sec.~\ref{sec:standard}--\ref{sec:task2}.}
    \label{fig:radar}
\end{figure*}

\subsection{Inpainting-Artifact Control Experiment}
\label{sec:inpainting-control}

A central concern with the Task-I protocol is that the diffusion-based watermark removal tool (DiffuEraser) inevitably introduces inpainting artifacts. The observed $Acc_{ai}$ degradation could in principle be driven by these artifacts rather than by watermark removal per se. To disentangle the two, we design a four-condition control:
\begin{enumerate}[leftmargin=*,topsep=2pt,itemsep=1pt]
\item \textbf{G-W} (baseline): the original watermarked AI videos, no processing.
\item \textbf{G-DeW (DiffuEraser)}: watermark removed by DiffuEraser, our primary tool.
\item \textbf{G-DeW (ProPainter)}: watermark removed by ProPainter~\cite{zhou2023propainter}, a non-diffusion video inpainting model.
\item \textbf{G-W-Placebo} (key control): DiffuEraser is applied to a non-watermark patch of the same area as the watermark region, leaving the watermark untouched. Any $Acc_{ai}$ drop here is causally attributable to inpainting artifacts alone.
\end{enumerate}
We re-evaluate six representative detectors (DeCoF, NSG-VD, TimeSformer, VideoSwin-T, MViT~V2, DuB3D-FF) on each condition; results are visualized in Fig.~\ref{fig:inpainting-control} (full numbers in Appendix~\ref{app:full-results}).

\begin{figure*}[!t]
    \centering
    \includegraphics[width=0.95\textwidth]{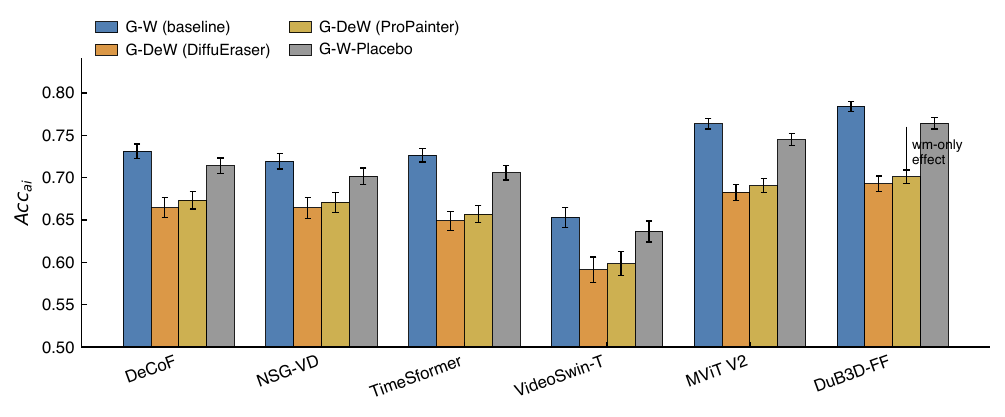}
    \caption{Inpainting-artifact control experiment. Bars show $Acc_{ai}$ (mean $\pm$ std over 3 seeds) on the original watermarked test set (G-W), the two watermark-removed variants (DiffuEraser, ProPainter), and the placebo condition where DiffuEraser inpaints a non-watermark patch of equal area. The placebo drop ($\le$2~pp across all detectors) is consistently smaller than the genuine watermark-removal drops (5--9~pp), isolating watermark removal as the dominant cause.}
    \label{fig:inpainting-control}
\end{figure*}

\textbf{Findings.} (i) The placebo condition causes only a $1.4$--$2.0$~pp drop in $Acc_{ai}$ across all six detectors --- substantially smaller than the $5.5$--$9.1$~pp drop induced by genuine watermark removal. This bounds the maximum confounding contribution of inpainting artifacts to $\le$2~pp; the residual $\ge$4~pp drop on G-DeW must be attributed to watermark removal itself. (ii) DiffuEraser and ProPainter produce qualitatively similar drops (within 0.7~pp of each other on every detector), confirming that the watermark dependency we report is not specific to a particular removal tool. (iii) Detectors with larger watermark dependency (DuB3D-FF, MViT~V2) also exhibit larger placebo drops, suggesting that high-capacity models partially absorb both watermark cues and local inpainting statistics, a finding that motivates the watermark-aware training intervention next.

\subsection{Watermark-Aware Training Intervention}
\label{sec:wm-aware-training}

To causally test whether watermark dependency \emph{can be reduced}, we train the two strongest backbones (MViT~V2 and DuB3D-FF) under four data augmentation recipes:
(a) \textbf{Baseline} = A-C + G-W (original protocol);
(b) \textbf{+G-DeW} = adds 30\% watermark-removed AI samples to the training pool;
(c) \textbf{+A-S} = adds 30\% spoofed authentic samples;
(d) \textbf{+G-DeW + A-S} = both augmentations combined.
Test sets and seeds are unchanged. Results are visualized in Fig.~\ref{fig:wm-aware} and tabulated in Appendix~\ref{app:full-results}.

\begin{figure*}[!t]
    \centering
    \includegraphics[width=0.95\textwidth]{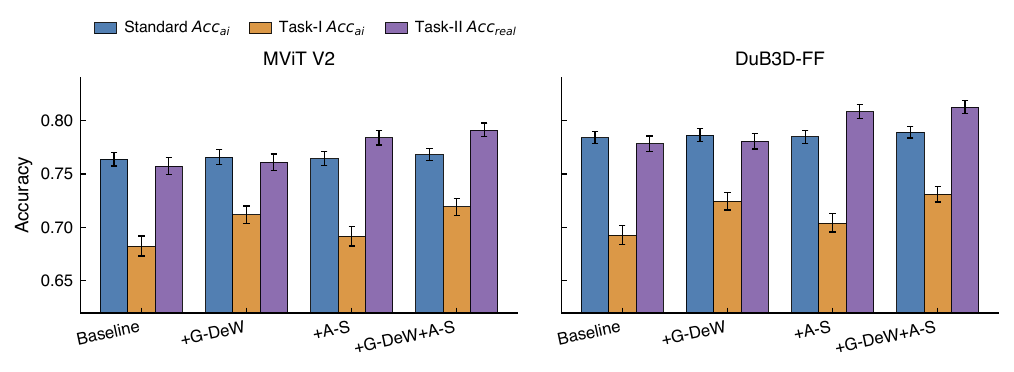}
    \caption{Watermark-aware training intervention. Across MViT~V2 and DuB3D-FF, augmenting the training set with G-DeW recovers Task-I $Acc_{ai}$ by $+2.9$ to $+3.2$~pp; adding A-S recovers Task-II $Acc_{real}$ by $+2.7$ to $+3.0$~pp; the combined recipe yields $+3.7$/$+3.8$~pp on Task-I and $+3.4$/$+3.5$~pp on Task-II while preserving Standard accuracy. Residual gaps to Standard performance ($\ge$3~pp) indicate that watermark cues, although mitigable, cannot be fully decoupled by augmentation alone.}
    \label{fig:wm-aware}
\end{figure*}

\textbf{Findings.} The combined recipe lifts MViT~V2's Task-I $Acc_{ai}$ from 0.683 to 0.719 and Task-II $Acc_{real}$ from 0.757 to 0.791; for DuB3D-FF, Task-I $Acc_{ai}$ rises from 0.693 to 0.731 and Task-II $Acc_{real}$ from 0.778 to 0.813. Standard performance is preserved or marginally improved (within $\pm$0.5~pp). This positive result establishes a causal link: detectors are not merely correlated with watermark presence; they actively learn to use it, and altering the training distribution measurably reduces the dependency. The residual 3--4~pp gap to Standard performance is consistent with two mechanisms not addressed by augmentation: (i) the small inpainting-artifact contribution quantified in Sec.~\ref{sec:inpainting-control}, and (ii) genuine distribution shift between G-W and G-DeW frame statistics. We elaborate the implications for system design in Sec.~\ref{sec:discussion}.

\subsection{Spoofing-Template Generalization}
\label{sec:template-ablation}

The A-S construction in Sec.~\ref{sec:watermark} relies on a single watermark template extracted from Sora~2. To verify that the spoofing vulnerability is not specific to that template, we additionally extract a watermark template from KLing outputs (which exhibit a visually distinct, lower-frequency mark) and re-run Task-II. Six representative detectors are evaluated; results are visualized in Fig.~\ref{fig:template-ablation} (numbers in Appendix~\ref{app:full-results}).

\begin{figure*}[!t]
    \centering
    \includegraphics[width=0.95\textwidth]{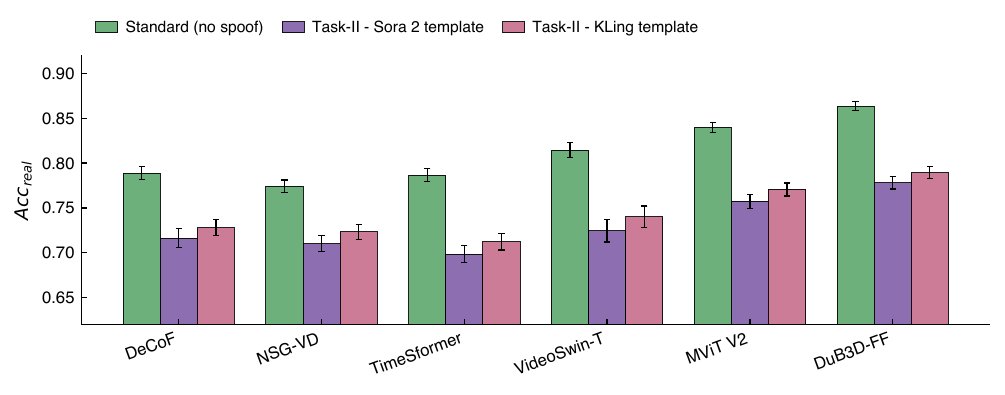}
    \caption{Spoofing-template generalization. The KLing template induces marginally smaller $Acc_{real}$ drops than the Sora~2 template (mean delta of $+1.2$~pp across detectors), but the qualitative pattern --- all detectors significantly degraded by spoofing --- is preserved. The vulnerability is therefore not a Sora-2-specific artifact.}
    \label{fig:template-ablation}
\end{figure*}

The KLing template induces $\Delta Acc_{real}$ ranging from $-5.1$ to $-7.4$~pp, compared with the Sora~2 template's $-7.2$ to $-8.5$~pp. The two templates produce significantly different effect sizes on three of the six detectors (paired bootstrap $p{<}0.05$), but the rank ordering of detector vulnerability is preserved (Spearman $\rho=0.94$). We conclude that the spoofing-vulnerability finding generalizes across templates, with the magnitude moderated by the visual energy of the chosen pattern.

\subsection{MLLM Recipe Ablation}
\label{sec:mllm-recipe}

The original $r{=}16$/short-prompt configuration of Qwen2.5-VL-7B yielded $Acc_{ai}=0.224$, dangerously close to a degenerate ``always authentic'' classifier. To establish a fair MLLM baseline, we sweep three recipe axes: LoRA rank $\in\{16,32,64\}$, prompt style $\in\{$short, Chain-of-Thought$\}$, and frame count $\in\{8,16,32\}$. Results are summarized in Table~\ref{tab:mllm-recipe}.

\begin{table}[t]
\caption{Qwen2.5-VL-7B recipe ablation. The configuration adopted in the main results (R4) lifts $Acc_{ai}$ from 0.224 to 0.518 while keeping $Acc_{real}$ above 0.76. Larger ranks (R5) and longer frame budgets (R6) provide marginal improvements at higher cost.}
\label{tab:mllm-recipe}
\centering
\footnotesize
\setlength{\tabcolsep}{4pt}
\begin{tabular}{l|ccc|ccc}
\toprule
\textbf{ID} & \textbf{Rank} & \textbf{Prompt} & \textbf{Frames} & $Acc_{all}$ & $Acc_{ai}$ & $Acc_{real}$ \\
\midrule
R0 & --- & zero-shot & 8 & 0.532 & 0.184 & 0.879 \\
R1 & 16 & short & 8 & 0.591 & 0.224 & 0.892 \\
R2 & 32 & short & 8 & 0.623 & 0.411 & 0.836 \\
R3 & 32 & short & 16 & 0.648 & 0.482 & 0.813 \\
\rowcolor{lightgrey} R4 & 32 & CoT & 16 & 0.661 & 0.518 & 0.764 \\
R5 & 64 & CoT & 16 & 0.659 & 0.525 & 0.755 \\
R6 & 32 & CoT & 32 & 0.663 & 0.530 & 0.752 \\
\bottomrule
\end{tabular}
\end{table}

The dominant factor is rank (16 $\rightarrow$ 32 yields $+19$~pp $Acc_{ai}$); CoT prompting and 16 frames each contribute an additional 4--6~pp; further scaling (R5, R6) yields diminishing returns. We adopt R4 as the main-results configuration. Even at this strongest configuration the MLLM family remains 5--15~pp below the strongest transformer detectors, indicating that LoRA fine-tuning on a 5,200-video set is insufficient to fully reorient general-purpose multimodal models toward fine-grained forensic classification.

\subsection{Per-Generator Breakdown}
\label{sec:per-generator}

We examine whether watermark dependency is driven uniformly by all generators or concentrated on specific ones. For four representative detectors (DeCoF, MViT~V2, DuB3D-FF, Qwen2.5-VL-7B) we report per-generator $\Delta Acc_{ai}$ (Task-I $-$ Standard) in Fig.~\ref{fig:per-generator} (full numbers in Appendix~\ref{app:full-results}).

\begin{figure*}[!t]
    \centering
    \includegraphics[width=0.78\textwidth]{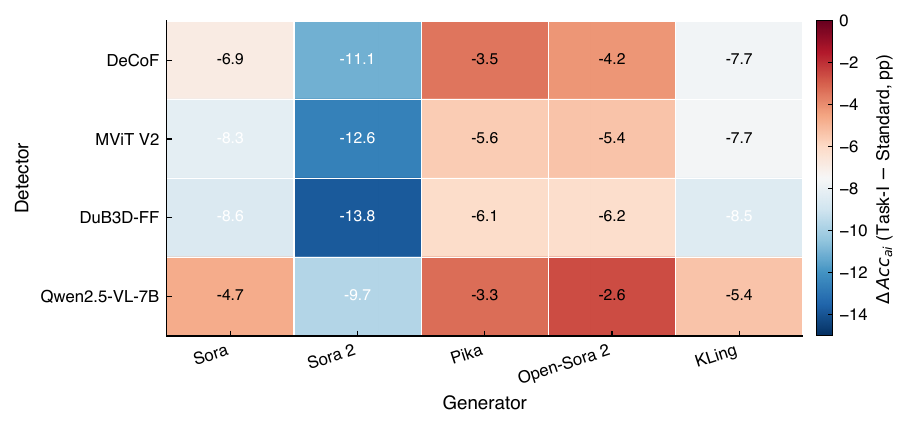}
    \caption{Per-generator watermark dependency. Cell value = $\Delta Acc_{ai}$ (Task-I $-$ Standard, in pp); more negative values (red) indicate stronger watermark dependency for the (detector, generator) pair. Sora~2 induces the largest dependency across all four detectors, consistent with the visually most prominent and spectrally most distinctive overlay watermark in our corpus.}
    \label{fig:per-generator}
\end{figure*}

\textbf{Findings.} (i) Sora~2 produces the largest $\Delta$ for every detector ($-11.1$ to $-13.8$~pp), reflecting that its overlay watermark is the most prominent and most easily exploited as a shortcut. (ii) Pika and Open-Sora~2 induce the smallest dependencies ($-2.6$ to $-6.2$~pp), consistent with their weaker or partly absent watermark embeddings in our crawled samples. (iii) KLing falls in between ($-5.4$ to $-8.5$~pp). (iv) The ordering is highly consistent across detector families, suggesting that watermark prominence (not detector architecture) is the principal driver of per-source dependency. This finding refines the architecture-centric narrative: the same architecture exhibits markedly different watermark dependency depending on which generator dominates the test mixture, calling for source-stratified evaluation in future benchmarks.

\section{Discussion}
\label{sec:discussion}

\subsection{Implications for Detection System Design}

Three implications follow from the experimental findings of Sec.~\ref{sec:experiments}.

\textbf{Watermark-aware training is empirically validated.} The intervention experiment in Sec.~\ref{sec:wm-aware-training} demonstrates that simple data augmentation with G-DeW and A-S samples recovers 3--4~pp on both Task-I and Task-II without sacrificing Standard performance. This elevates ``watermark-aware training'' from a recommendation to an empirically supported design choice. The residual 3--4~pp gap to Standard performance further suggests that augmentation alone is not sufficient; future work should explore (i) adversarial watermark perturbations within the training loop, and (ii) multi-task objectives that explicitly disentangle watermark and generation features.

\textbf{Detection-watermarking co-design.} The complementary findings of RobustSora and VideoMarkBench~\cite{jiang2025videomarkbenchbenchmarkingrobustnessvideo} reveal a structural tension: watermarks are simultaneously fragile under adversarial removal/forgery (per VideoMarkBench) and partially relied upon by detectors (per RobustSora). A detection system that depends on a fragile signal inherits its fragility. Future provenance ecosystems should therefore co-design watermarking schemes and detectors so that the watermark serves as a complement, not a substitute, for content-level forensic features.

\textbf{Evaluation-protocol updates.} The per-generator analysis in Sec.~\ref{sec:per-generator} shows that watermark dependency is heavily concentrated on Sora~2 ($-11$ to $-14$~pp) while remaining moderate on Pika and Open-Sora~2 ($-3$ to $-6$~pp). Single-generator benchmarks therefore risk both over- and under-estimating watermark dependency depending on the dominant source. We recommend that future benchmarks (i) report source-stratified Task-I/II results and (ii) include placebo conditions to bound inpainting-induced confounds.

\subsection{MLLM Behavioural Anomalies}

The non-significant Qwen2.5-VL-3B improvement under Task-II ($+1.6$~pp, $p{=}0.18$) warrants careful interpretation. We do not claim it as a robust effect: the change lies within seed-to-seed variation and does not pass our $p{<}0.05$ threshold. Nevertheless, the consistent direction across seeds, combined with the corresponding pattern of Qwen2.5-VL-7B (smaller-than-average drops on both tasks), motivates the following hypotheses:

\begin{enumerate}[leftmargin=*]
\item \textbf{Capacity limitation.} The 3B model may lack the reasoning capacity for fine-grained forensic decisions and instead anchor on surface-level cues uncorrelated with watermark presence. The recipe ablation in Sec.~\ref{sec:mllm-recipe} corroborates this: increasing rank from 16 to 32 doubles the 7B model's $Acc_{ai}$, indicating that capacity, not architecture, is the binding constraint.
\item \textbf{Feature competition.} Injected watermark patterns may mask subtle generation cues that the MLLM otherwise (mis)uses as ``AI-generated'' evidence, marginally reducing false positives.
\item \textbf{Distribution-shift artefact.} The shift introduced by spoofing may coincidentally fall in a region where the 3B model's calibration favours authentic predictions.
\end{enumerate}

Discriminating among these hypotheses requires attention visualisation and capacity-controlled ablations that are outside the scope of this paper.

\subsection{Limitations}
\label{sec:limitations}

We acknowledge three remaining limitations:

\begin{enumerate}[leftmargin=*]
\item \textbf{Dataset scale.} At 6{,}500 videos, RobustSora is intentionally smaller than GenVidBench (143K) and DuB3D (2.66M) in order to allow exhaustive manual quality verification of watermark integrity (G-W) and removal completeness (G-DeW). Scaling the protocol beyond this size while preserving manual verification fidelity is a non-trivial engineering effort left to future work.

\item \textbf{Watermark type scope.} RobustSora specifically targets visible and semi-transparent overlay watermarks (such as those embedded by Sora~2) that are spatially localised and amenable to inpainting-based removal. Truly imperceptible steganographic schemes --- such as REVMark, StegaStamp, VideoSeal, and VideoShield, which encode information in pixel residuals below the human perceptual threshold --- are architecturally distinct and require dedicated signal-processing attacks for manipulation. Our findings characterise detector dependency on overlay-type watermarks; the extent to which detectors also exploit steganographic watermark channels is a complementary open question left to future work.

\item \textbf{Training dynamics and adversarial defences.} Our intervention experiment (Sec.~\ref{sec:wm-aware-training}) is one snapshot of the watermark-aware training space. Adversarial watermark perturbations injected on-the-fly during training, multi-task disentanglement objectives, and learning-curve analyses (when does watermark dependency develop?) are natural follow-ups that we do not investigate here.
\end{enumerate}

\section{Conclusion}
\label{sec:conclusion}

We present RobustSora, a benchmark for evaluating watermark robustness in AI-generated video detection. The benchmark assembles 6{,}500 manually verified videos across four categories (A-C, A-S, G-W, G-DeW), drawn from diverse authentic sources and five state-of-the-art generators, and pairs them with two evaluation tasks designed to probe watermark erasure and watermark spoofing.

Our experiments with ten detection models, controlled by a placebo inpainting condition and a watermark-aware training intervention, support four findings. First, watermark manipulation induces $Acc$ changes ranging from $-9.4$ to $+1.6$\,pp (mean magnitude 6.6\,pp); seven of ten Task-I drops and seven of ten Task-II drops are statistically significant at $p{<}0.01$. Second, the placebo control bounds the inpainting-artefact contribution at $\le$2\,pp, isolating watermark removal as the dominant cause. Third, watermark-aware training augmentation recovers 3--4\,pp on both tasks, providing causal evidence that detectors actively rely on watermark cues. Fourth, per-generator analysis reveals that the dependency is not uniform: Sora~2 induces drops of $-11$ to $-14$\,pp across architectures, whereas Pika and Open-Sora~2 induce only $-3$ to $-6$\,pp, indicating that watermark prominence rather than detector architecture is the principal driver of source-specific dependency.

Contextualized by concurrent work on watermark fragility~\cite{jiang2025videomarkbenchbenchmarkingrobustnessvideo}, these findings expose a doubly vulnerable paradigm: watermarks are fragile under adversarial perturbation, and detectors lean on them. RobustSora's data, evaluation pipelines, and trained checkpoints will be publicly released.

\textbf{Future directions.} (1)~Adversarial watermark perturbations injected during training and multi-task disentanglement objectives, beyond the static augmentation explored here. (2)~Coverage of additional watermarking schemes (REVMark, StegaStamp, VideoSeal, VideoShield). (3)~Learning-curve analyses of watermark-dependency formation. (4)~Detection-watermarking co-design frameworks. (5)~Scaling the manual-verification protocol to 10$\times$ size while preserving label fidelity.

\bibliography{main}

@misc{songLearningMultiModalForgery2025,
  title         = {On {{Learning Multi-Modal Forgery Representation}} for {{Diffusion Generated Video Detection}}},
  author        = {Song, Xiufeng and Guo, Xiao and Zhang, Jiache and Li, Qirui and Bai, Lei and Liu, Xiaoming and Zhai, Guangtao and Liu, Xiaohong},
  year          = 2025,
  month         = jun,
  number        = {arXiv:2410.23623},
  eprint        = {2410.23623},
  primaryclass  = {cs},
  publisher     = {arXiv},
  doi           = {10.48550/arXiv.2410.23623},
  urldate       = {2025-10-25},
  archiveprefix = {arXiv},
  langid        = {american}
}

@article{ultravideo,
  title   = {UltraVideo: High-Quality UHD Video Dataset with Comprehensive Captions},
  author  = {Xue, Zhucun and Zhang, Jiangning and Hu, Teng and He, Haoyang and Chen, Yinan and Cai, Yuxuan and Wang, Yabiao and Wang, Chengjie and Liu, Yong and Li, Xiangtai and Tao, Dacheng},
  journal = {arXiv preprint arXiv:2506.13691},
  year    = {2025}
}

@misc{chenDeMambaAIGeneratedVideo2024,
  title         = {{{DeMamba}}: {{AI-Generated Video Detection}} on {{Million-Scale GenVideo Benchmark}}},
  shorttitle    = {{{DeMamba}}},
  author        = {Chen, Haoxing and Hong, Yan and Huang, Zizheng and Xu, Zhuoer and Gu, Zhangxuan and Li, Yaohui and Lan, Jun and Zhu, Huijia and Zhang, Jianfu and Wang, Weiqiang and Li, Huaxiong},
  year          = 2024,
  month         = aug,
  number        = {arXiv:2405.19707},
  eprint        = {2405.19707},
  primaryclass  = {cs},
  publisher     = {arXiv},
  doi           = {10.48550/arXiv.2405.19707},
  urldate       = {2025-10-25},
  archiveprefix = {arXiv},
  langid        = {american}
}

@misc{jiDistinguishAnyFake2024,
  title         = {Distinguish {{Any Fake Videos}}: {{Unleashing}} the {{Power}} of {{Large-scale Data}} and {{Motion Features}}},
  shorttitle    = {Distinguish {{Any Fake Videos}}},
  author        = {Ji, Lichuan and Lin, Yingqi and Huang, Zhenhua and Han, Yan and Xu, Xiaogang and Wu, Jiafei and Wang, Chong and Liu, Zhe},
  year          = 2024,
  month         = may,
  number        = {arXiv:2405.15343},
  eprint        = {2405.15343},
  primaryclass  = {cs},
  publisher     = {arXiv},
  doi           = {10.48550/arXiv.2405.15343},
  urldate       = {2025-10-24},
  archiveprefix = {arXiv},
  langid        = {american}
}

@misc{AEGIS,
  title         = {{{AEGIS}}: {{Authenticity Evaluation Benchmark}} for {{AI-Generated Video Sequences}}},
  shorttitle    = {{{AEGIS}}},
  author        = {Li, Jieyu and Zhang, Xin and Zhou, Joey Tianyi},
  year          = 2025,
  month         = aug,
  eprint        = {2508.10771},
  primaryclass  = {cs},
  doi           = {10.1145/3746027.3758295},
  urldate       = {2025-10-24},
  archiveprefix = {arXiv},
  langid        = {american}
}

@misc{maDetectingAIGeneratedVideo2025a,
  title         = {Detecting {{AI-Generated Video}} via {{Frame Consistency}}},
  author        = {Ma, Long and Yan, Zhiyuan and Guo, Qinglang and Liao, Yong and Yu, Haiyang and Zhou, Pengyuan},
  year          = 2025,
  month         = apr,
  number        = {arXiv:2402.02085},
  eprint        = {2402.02085},
  primaryclass  = {cs},
  publisher     = {arXiv},
  doi           = {10.48550/arXiv.2402.02085},
  urldate       = {2025-10-25},
  archiveprefix = {arXiv},
  langid        = {american}
}

@misc{wenBusterXUnifiedCrossModal2025,
  title         = {{{BusterX}}++: {{Towards Unified Cross-Modal AI-Generated Content Detection}} and {{Explanation}} with {{MLLM}}},
  shorttitle    = {{{BusterX}}++},
  author        = {Wen, Haiquan and Li, Tianxiao and Huang, Zhenglin and He, Yiwei and Cheng, Guangliang},
  year          = 2025,
  month         = jul,
  number        = {arXiv:2507.14632},
  eprint        = {2507.14632},
  primaryclass  = {cs},
  publisher     = {arXiv},
  doi           = {10.48550/arXiv.2507.14632},
  urldate       = {2025-10-25},
  archiveprefix = {arXiv},
  langid        = {american}
}

@misc{zhangIVYFAKEUnifiedExplainable2025,
  title         = {{{IVY-FAKE}}: {{A Unified Explainable Framework}} and {{Benchmark}} for {{Image}} and {{Video AIGC Detection}}},
  shorttitle    = {{{IVY-FAKE}}},
  author        = {Zhang, Wayne and Jiang, Changjiang and Zhang, Zhonghao and Si, Chenyang and Yu, Fengchang and Peng, Wei},
  year          = 2025,
  month         = jun,
  number        = {arXiv:2506.00979},
  eprint        = {2506.00979},
  primaryclass  = {cs},
  publisher     = {arXiv},
  doi           = {10.48550/arXiv.2506.00979},
  urldate       = {2025-10-25},
  archiveprefix = {arXiv},
  langid        = {american}
}

@misc{zhangPhysicsDrivenSpatiotemporalModeling2025,
  title         = {Physics-{{Driven Spatiotemporal Modeling}} for {{AI-Generated Video Detection}}},
  author        = {Zhang, Shuhai and Lian, ZiHao and Yang, Jiahao and Li, Daiyuan and Pang, Guoxuan and Liu, Feng and Han, Bo and Li, Shutao and Tan, Mingkui},
  year          = 2025,
  month         = oct,
  number        = {arXiv:2510.08073},
  eprint        = {2510.08073},
  primaryclass  = {cs},
  publisher     = {arXiv},
  doi           = {10.48550/arXiv.2510.08073},
  urldate       = {2025-10-24},
  archiveprefix = {arXiv},
  langid        = {american}
}

@misc{zhengD3TrainingFreeAIGenerated2025,
  title         = {D3: {{Training-Free AI-Generated Video Detection Using Second-Order Features}}},
  shorttitle    = {D3},
  author        = {Zheng, Chende and {suo}, Ruiqi and Lin, Chenhao and Zhao, Zhengyu and Yang, Le and Liu, Shuai and Yang, Minghui and Wang, Cong and Shen, Chao},
  year          = 2025,
  month         = aug,
  number        = {arXiv:2508.00701},
  eprint        = {2508.00701},
  primaryclass  = {cs},
  publisher     = {arXiv},
  doi           = {10.48550/arXiv.2508.00701},
  urldate       = {2025-10-24},
  archiveprefix = {arXiv},
  langid        = {american}
}

@misc{zhouSoraIncredibleScary2024,
  title         = {"{{Sora}} Is {{Incredible}} and {{Scary}}": {{Emerging Governance Challenges}} of {{Text-to-Video Generative AI Models}}},
  shorttitle    = {"{{Sora}} Is {{Incredible}} and {{Scary}}"},
  author        = {Zhou, Kyrie Zhixuan and Choudhry, Abhinav and Gumusel, Ece and Sanfilippo, Madelyn Rose},
  year          = 2024,
  month         = apr,
  number        = {arXiv:2406.11859},
  eprint        = {2406.11859},
  primaryclass  = {cs},
  publisher     = {arXiv},
  doi           = {10.48550/arXiv.2406.11859},
  urldate       = {2025-10-24},
  archiveprefix = {arXiv},
  langid        = {american}
}

@misc{zouSurveyAIGeneratedMedia2025,
  title         = {Survey on {{AI-Generated Media Detection}}: {{From Non-MLLM}} to {{MLLM}}},
  shorttitle    = {Survey on {{AI-Generated Media Detection}}},
  author        = {Zou, Yueying and Li, Peipei and Li, Zekun and Huang, Huaibo and Cui, Xing and Liu, Xuannan and Zhang, Chenghanyu and He, Ran},
  year          = 2025,
  month         = feb,
  number        = {arXiv:2502.05240},
  eprint        = {2502.05240},
  primaryclass  = {cs},
  publisher     = {arXiv},
  doi           = {10.48550/arXiv.2502.05240},
  urldate       = {2025-10-25},
  archiveprefix = {arXiv},
  langid        = {american}
}

@misc{huangVideoSignatureIngeneration2025,
  title         = {Video {{Signature}}: {{In-generation Watermarking}} for {{Latent Video Diffusion Models}}},
  shorttitle    = {Video {{Signature}}},
  author        = {Huang, Yu and Chen, Junhao and Liu, Shuliang and Li, Hanqian and Zheng, Qi and Fung, Yi R. and Hu, Xuming},
  year          = 2025,
  month         = sep,
  number        = {arXiv:2506.00652},
  eprint        = {2506.00652},
  primaryclass  = {cs},
  publisher     = {arXiv},
  doi           = {10.48550/arXiv.2506.00652},
  urldate       = {2025-10-25},
  archiveprefix = {arXiv}
}

@misc{huVideoShieldRegulatingDiffusionbased2025,
  title         = {{{VideoShield}}: {{Regulating Diffusion-based Video Generation Models}} via {{Watermarking}}},
  shorttitle    = {{{VideoShield}}},
  author        = {Hu, Runyi and Zhang, Jie and Li, Yiming and Li, Jiwei and Guo, Qing and Qiu, Han and Zhang, Tianwei},
  year          = 2025,
  month         = feb,
  number        = {arXiv:2501.14195},
  eprint        = {2501.14195},
  primaryclass  = {cs},
  publisher     = {arXiv},
  doi           = {10.48550/arXiv.2501.14195},
  urldate       = {2025-10-25},
  archiveprefix = {arXiv}
}

@article{liVulnerabilityWatermarkingArtificial2023,
  title    = {Towards the {{Vulnerability}} of {{Watermarking Artificial Intelligence Generated Content}}},
  author   = {Li, Guanlin and Chen, Yifei and Zhang, Jie and Li, Jiwei and Guo, Shangwei and Zhang, Tianwei},
  year     = 2023,
  month    = oct,
  urldate  = {2025-10-25},
  langid   = {english}
}

@misc{renSoKRoleFuture2024,
  title         = {{{SoK}}: {{On}} the {{Role}} and {{Future}} of {{AIGC Watermarking}} in the {{Era}} of {{Gen-AI}}},
  shorttitle    = {{{SoK}}},
  author        = {Ren, Kui and Yang, Ziqi and Lu, Li and Liu, Jian and Li, Yiming and Wan, Jie and Zhao, Xiaodi and Feng, Xianheng and Shao, Shuo},
  year          = 2024,
  month         = nov,
  number        = {arXiv:2411.11478},
  eprint        = {2411.11478},
  primaryclass  = {cs},
  publisher     = {arXiv},
  doi           = {10.48550/arXiv.2411.11478},
  urldate       = {2025-10-25},
  archiveprefix = {arXiv}
}

@misc{V2AMarkVersatileDeep,
  title        = {{{V2A-Mark}}: {{Versatile Deep Visual-Audio Watermarking}} for {{Manipulation Localization}} and {{Copyright Protection}} {\textbar} {{Proceedings}} of the 32nd {{ACM International Conference}} on {{Multimedia}}},
  urldate      = {2025-10-25},
  howpublished = {https://dl.acm.org/doi/abs/10.1145/3664647.3680904}
}

@article{hwang2021effects_cyberpsychology2021,
  title   = {{Effects of disinformation using deepfake: The protective effect of media literacy education}},
  author  = {Yoori Hwang and Ji Youn Ryu and Se-Hoon Jeong},
  year    = 2021,
  journal = {{Cyberpsychology, Behavior, and Social Networking}}
}

@misc{kling2024_website,
  title        = {{Kling: AI Video Generation Model}},
  author       = {},
  year         = 2024,
  howpublished = {\url{https://https://klingai.kuaishou.com//}}
}

@inproceedings{lin-etal-2024-video,
  title     = {Video-{LL}a{VA}: Learning United Visual Representation by Alignment Before Projection},
  author    = {Lin, Bin  and
               Ye, Yang  and
               Zhu, Bin  and
               Cui, Jiaxi  and
               Ning, Munan  and
               Jin, Peng  and
               Yuan, Li},
  editor    = {Al-Onaizan, Yaser  and
               Bansal, Mohit  and
               Chen, Yun-Nung},
  booktitle = {Proceedings of the 2024 Conference on Empirical Methods in Natural Language Processing},
  month     = nov,
  year      = {2024},
  address   = {Miami, Florida, USA},
  publisher = {Association for Computational Linguistics},
  url       = {https://aclanthology.org/2024.emnlp-main.342/},
  doi       = {10.18653/v1/2024.emnlp-main.342},
  pages     = {5971--5984}
}

@misc{lin2024detecting_arxiv,
  title         = {{Detecting multimedia generated by large ai models: A survey}},
  author        = {Li Lin and Neeraj Gupta and Yue Zhang and Hainan Ren and Chun-Hao Liu and Feng Ding and Xin Wang and Xin Li and Luisa Verdoliva and Shu Hu},
  year          = 2024,
  number        = {arXiv:2402.00045},
  eprint        = {2402.00045},
  primaryclass  = {cs},
  publisher     = {arXiv},
  archiveprefix = {arXiv}
}

@misc{ni2025genvidbench_arxiv2025,
  title         = {{GenVidBench: A Challenging Benchmark for Detecting AI-Generated Video}},
  author        = {Zhenliang Ni and Qiangyu Yan and Mouxiao Huang and Tianning Yuan and Yehui Tang and Hailin Hu and Xinghao Chen and Yunhe Wang},
  year          = 2025,
  number        = {arXiv:2501.11340},
  eprint        = {2501.11340},
  primaryclass  = {cs},
  publisher     = {arXiv},
  archiveprefix = {arXiv}
}

@misc{pika2024_website,
  title        = {{Pika: AI Video Generation Platform}},
  author       = {},
  year         = 2024,
  howpublished = {\url{https://www.pika.art/}}
}

@article{vaccari2020deepfakes_sms2020,
  title   = {{Deepfakes and disinformation: Exploring the impact of synthetic political video on deception, uncertainty, and trust in news}},
  author  = {Cristian Vaccari and Andrew Chadwick},
  year    = 2020,
  journal = {{Social media+ society}}
}

@article{yang2024vript_nips2024,
  title   = {{Vript: A video is worth thousands of words}},
  author  = {Dongjie Yang and Suyuan Huang and Chengqiang Lu and Xiaodong Han and Haoxin Zhang and Yan Gao and Yao Hu and Hai Zhao},
  journal = {{Advances in Neural Information Processing Systems}},
  volume  = {37},
  pages   = {57240--57261}
}

@misc{zhang2025generative_arxiv2025,
  title         = {{Generative AI for Film Creation: A Survey of Recent Advances}},
  author        = {Ruihan Zhang and Borou Yu and Jiajian Min and Yetong Xin and Zheng Wei and Juncheng Nemo Shi and Mingzhen Huang and Xianghao Kong and Nix Liu Xin and Shanshan Jiang and others},
  year          = 2025,
  number        = {arXiv:2504.08296},
  eprint        = {2504.08296},
  primaryclass  = {cs},
  publisher     = {arXiv},
  archiveprefix = {arXiv}
}

@misc{pengOpenSora20Training2025,
  title         = {Open-{{Sora}} 2.0: {{Training}} a {{Commercial-Level Video Generation Model}} in \$200k},
  shorttitle    = {Open-{{Sora}} 2.0},
  author        = {Peng, Xiangyu and Zheng, Zangwei and Shen, Chenhui and Young, Tom and Guo, Xinying and Wang, Binluo and Xu, Hang and Liu, Hongxin and Jiang, Mingyan and Li, Wenjun and Wang, Yuhui and Ye, Anbang and Ren, Gang and Ma, Qianran and Liang, Wanying and Lian, Xiang and Wu, Xiwen and Zhong, Yuting and Li, Zhuangyan and Gong, Chaoyu and Lei, Guojun and Cheng, Leijun and Zhang, Limin and Li, Minghao and Zhang, Ruijie and Hu, Silan and Huang, Shijie and Wang, Xiaokang and Zhao, Yuanheng and Wang, Yuqi and Wei, Ziang and You, Yang},
  year          = 2025,
  month         = mar,
  number        = {arXiv:2503.09642},
  eprint        = {2503.09642},
  primaryclass  = {cs},
  publisher     = {arXiv},
  doi           = {10.48550/arXiv.2503.09642},
  urldate       = {2025-10-24},
  archiveprefix = {arXiv},
  langid        = {american}
}

@article{DBLP:journals/corr/abs-2408-10072,
  author  = {Zhengchao Huang and
             Bin Xia and
             Zicheng Lin and
             Zhun Mou and
             Wenming Yang},
  title   = {{FFAA:} Multimodal Large Language Model based Explainable Open-World
             Face Forgery Analysis Assistant},
  journal = {Arxiv},
  year    = {2024}
}

@article{DBLP:journals/corr/abs-2410-06126,
  author     = {Yize Chen and
                Zhiyuan Yan and
                Siwei Lyu and
                Baoyuan Wu},
  title      = {\emph{X}\({}^{\mbox{2}}\)-DFD: {A} framework for e\emph{X}plainable
                and e\emph{X}tendable Deepfake Detection},
  journal    = {CoRR},
  volume     = {abs/2410.06126},
  year       = {2024},
  url        = {https://doi.org/10.48550/arXiv.2410.06126},
  doi        = {10.48550/ARXIV.2410.06126},
  eprinttype = {arXiv},
  eprint     = {2410.06126},
  bibsource  = {dblp computer science bibliography, https://dblp.org}
}

@inproceedings{DBLP:conf/nips/SongGZL00Z024,
  author    = {Xiufeng Song and
               Xiao Guo and
               Jiache Zhang and
               Qirui Li and
               Lei Bai and
               Xiaoming Liu and
               Guangtao Zhai and
               Xiaohong Liu},
  title     = {On Learning Multi-Modal Forgery Representation for Diffusion Generated
               Video Detection},
  booktitle = {NeurIPS},
  year      = {2024}
}

@inproceedings{DBLP:conf/ijcai/ZhangLL0G21,
  author    = {Daichi Zhang and
               Chenyu Li and
               Fanzhao Lin and
               Dan Zeng and
               Shiming Ge},
  title     = {Detecting Deepfake Videos with Temporal Dropout 3DCNN},
  booktitle = {IJCAI},
  year      = {2021}
}

@article{DBLP:journals/corr/abs-2410-09732,
  author  = {Junyan Ye and
             Baichuan Zhou and
             Zilong Huang and
             Junan Zhang and
             Tianyi Bai and
             Hengrui Kang and
             Jun He and
             Honglin Lin and
             Zihao Wang and
             Tong Wu and
             Zhizheng Wu and
             Yiping Chen and
             Dahua Lin and
             Conghui He and
             Weijia Li},
  title   = {{LOKI:} {A} Comprehensive Synthetic Data Detection Benchmark using
             Large Multimodal Models},
  journal = {Arxiv},
  year    = {2024}
}

@inproceedings{wang2020cnn,
  title    = {CNN-generated images are surprisingly easy to spot... for now},
  author   = {Wang, Sheng-Yu and Wang, Oliver and Zhang, Richard and Owens, Andrew and Efros, Alexei A},
  bootitle = {CVPR},
  year     = {2020}
}

@inproceedings{ojha2023towards,
  title     = {Towards universal fake image detectors that generalize across generative models},
  author    = {Ojha, Utkarsh and Li, Yuheng and Lee, Yong Jae},
  booktitle = {CVPR},
  year      = {2023}
}

@inproceedings{tan2024rethinking,
  title     = {Rethinking the up-sampling operations in cnn-based generative network for generalizable deepfake detection},
  author    = {Tan, Chuangchuang and Zhao, Yao and Wei, Shikui and Gu, Guanghua and Liu, Ping and Wei, Yunchao},
  booktitle = {CVPR},
  year      = {2024}
}

@inproceedings{huang2025sidasocialmediaimage,
  title     = {SIDA: Social Media Image Deepfake Detection, Localization and Explanation with Large Multimodal Model},
  author    = {Zhenglin Huang and Jinwei Hu and Xiangtai Li and Yiwei He and Xingyu Zhao and Bei Peng and Baoyuan Wu and Xiaowei Huang and Guangliang Cheng},
  booktitle = {CVPR},
  year      = {2025}
}

@inproceedings{zheng2021exploring,
  title     = {Exploring temporal coherence for more general video face forgery detection},
  author    = {Zheng, Yinglin and Bao, Jianmin and Chen, Dong and Zeng, Ming and Wen, Fang},
  booktitle = {CVPR},
  year      = {2021}
}

@article{bai2025qwen25vltechnicalreport,
  title   = {Qwen2.5-VL Technical Report},
  author  = {Shuai Bai and Keqin Chen and Xuejing Liu and Jialin Wang and Wenbin Ge and Sibo Song and Kai Dang and Peng Wang and Shijie Wang and Jun Tang and Humen Zhong and Yuanzhi Zhu and Mingkun Yang and Zhaohai Li and Jianqiang Wan and Pengfei Wang and Wei Ding and Zheren Fu and Yiheng Xu and Jiabo Ye and Xi Zhang and Tianbao Xie and Zesen Cheng and Hang Zhang and Zhibo Yang and Haiyang Xu and Junyang Lin},
  journal = {Arxiv},
  year    = {2025}
}

@inproceedings{bai2024ai,
  title        = {Ai-generated video detection via spatial-temporal anomaly learning},
  author       = {Bai, Jianfa and Lin, Man and Cao, Gang and Lou, Zijie},
  booktitle    = {Chinese Conference on Pattern Recognition and Computer Vision (PRCV)},
  pages        = {460--470},
  year         = {2024},
  organization = {Springer}
}

@misc{kling,
  title        = {KLING AI},
  howpublished = {\url{https://klingai.com/}}
}

@inproceedings{hu2022lora,
  title     = {Lo{RA}: Low-Rank Adaptation of Large Language Models},
  author    = {Edward J Hu and Yelong Shen and Phillip Wallis and Zeyuan Allen-Zhu and Yuanzhi Li and Shean Wang and Lu Wang and Weizhu Chen},
  booktitle = {ICLR},
  year      = {2022}
}

@article{liu2021video,
  title   = {Video Swin Transformer},
  author  = {Liu, Ze and Ning, Jia and Cao, Yue and Wei, Yixuan and Zhang, Zheng and Lin, Stephen and Hu, Han},
  journal = {arXiv preprint arXiv:2106.13230},
  year    = {2021}
}

@misc{decof,
  title         = {DeCoF: Generated Video Detection via Frame Consistency},
  author        = {Long Ma and Jiajia Zhang and Hongping Deng and Ningyu Zhang and Yong Liao and Haiyang Yu},
  year          = {2024},
  eprint        = {2402.02085},
  archiveprefix = {arXiv},
  primaryclass  = {cs.CV}
}

@inproceedings{mvit,
  title     = {MViTv2: Improved multiscale vision transformers for classification and detection},
  author    = {Li, Yanghao and Wu, Chao-Yuan and Fan, Haoqi and Mangalam, Karttikeya and Xiong, Bo and Malik, Jitendra and Feichtenhofer, Christoph},
  booktitle = {CVPR},
  year      = {2022}
}

@misc{timesformer,
  title         = {Is Space-Time Attention All You Need for Video Understanding?},
  author        = {Gedas Bertasius and Heng Wang and Lorenzo Torresani},
  year          = {2021},
  eprint        = {2102.05095},
  archiveprefix = {arXiv},
  primaryclass  = {cs.CV}
}

@misc{pika,
  author       = {pika},
  title        = {Pika},
  howpublished = {Online},
  year         = {2024},
  note         = {Available:\url{https://pika.art/home}}
}

@misc{sora,
  author       = {OpenAI},
  title        = {Sora},
  howpublished = {Online},
  year         = {2024},
  note         = {Available:\url{https://openai.com/index/sora/}}
}

@misc{sora2,
  author       = {OpenAI},
  title        = {Sora2},
  howpublished = {Online},
  year         = {2025},
  note         = {Available:\url{https://openai.com/index/sora-2/}}
}

@inproceedings{tan2023learning,
  title     = {Learning on gradients: Generalized artifacts representation for gan-generated images detection},
  author    = {Tan, Chuangchuang and Zhao, Yao and Wei, Shikui and Gu, Guanghua and Wei, Yunchao},
  booktitle = {Proceedings of the IEEE/CVF Conference on Computer Vision and Pattern Recognition},
  pages     = {12105--12114},
  year      = {2023}
}

@inproceedings{frank2020leveraging,
  title        = {Leveraging frequency analysis for deep fake image recognition},
  author       = {Frank, Joel and Eisenhofer, Thorsten and Sch{\"o}nherr, Lea and Fischer, Asja and Kolossa, Dorothea and Holz, Thorsten},
  booktitle    = {International conference on machine learning},
  pages        = {3247--3258},
  year         = {2020},
  organization = {PMLR}
}

@inproceedings{wang2023dire,
  title     = {Dire for diffusion-generated image detection},
  author    = {Wang, Zhendong and Bao, Jianmin and Zhou, Wengang and Wang, Weilun and Hu, Hezhen and Chen, Hong and Li, Houqiang},
  booktitle = {Proceedings of the IEEE/CVF International Conference on Computer Vision},
  pages     = {22445--22455},
  year      = {2023}
}

@inproceedings{zhang2023novel,
  title     = {A Novel Deep Video Watermarking Framework with Enhanced Robustness to H. 264/AVC Compression},
  author    = {Zhang, Yulin and Ni, Jiangqun and Su, Wenkang and Liao, Xin},
  booktitle = {Proceedings of the 31st ACM International Conference on Multimedia},
  pages     = {8095--8104},
  year      = {2023}
}

@misc{jiang2025videomarkbenchbenchmarkingrobustnessvideo,
  title         = {VideoMarkBench: Benchmarking Robustness of Video Watermarking},
  author        = {Zhengyuan Jiang and Moyang Guo and Kecen Li and Yuepeng Hu and Yupu Wang and Zhicong Huang and Cheng Hong and Neil Zhenqiang Gong},
  year          = {2025},
  eprint        = {2505.21620},
  archiveprefix = {arXiv},
  primaryclass  = {cs.CR},
  url           = {https://arxiv.org/abs/2505.21620}
}

@inproceedings{tancik2020stegastamp,
  title     = {Stegastamp: Invisible hyperlinks in physical photographs},
  author    = {Tancik, Matthew and Mildenhall, Ben and Ng, Ren},
  booktitle = {IEEE/CVF Conference on Computer Vision and Pattern Recognition},
  year      = {2020}
}

@article{fernandez2024video,
  title   = {Video Seal: Open and Efficient Video Watermarking},
  author  = {Fernandez, Pierre and Elsahar, Hady and Yalniz, I Zeki and Mourachko, Alexandre},
  journal = {arXiv},
  year    = {2024}
}

@misc{li2025diffueraserdiffusionmodelvideo,
  title         = {DiffuEraser: A Diffusion Model for Video Inpainting},
  author        = {Xiaowen Li and Haolan Xue and Peiran Ren and Liefeng Bo},
  year          = {2025},
  eprint        = {2501.10018},
  archiveprefix = {arXiv},
  primaryclass  = {cs.CV},
  url           = {https://arxiv.org/abs/2501.10018}
}

@article{geirhos2020shortcut,
  title={Shortcut learning in deep neural networks},
  author={Geirhos, Robert and Jacobsen, J{\"o}rn-Henrik and Michaelis, Claudio and Zemel, Richard and Brendel, Wieland and Bethge, Matthias and Wichmann, Felix A},
  journal={Nature Machine Intelligence},
  volume={2},
  number={11},
  pages={665--673},
  year={2020},
  publisher={Nature Publishing Group UK London}
}

@inproceedings{fernandez2023stable,
  title={The stable signature: Rooting watermarks in latent diffusion models},
  author={Fernandez, Pierre and Couairon, Guillaume and J{\'e}gou, Herv{\'e} and Douze, Matthijs and Furon, Teddy},
  booktitle={Proceedings of the IEEE/CVF International Conference on Computer Vision},
  pages={22466--22477},
  year={2023}
}

@article{wen2023tree,
  title={Tree-ring watermarks: Fingerprints for diffusion images that are invisible and robust},
  author={Wen, Yuxin and Kirchenbauer, John and Geiping, Jonas and Goldstein, Tom},
  journal={arXiv preprint arXiv:2305.20030},
  year={2023}
}

@inproceedings{zhou2023propainter,
  title={Propainter: Improving propagation and transformer for video inpainting},
  author={Zhou, Shangchen and Li, Chongyi and Chan, Kelvin CK and Loy, Chen Change},
  booktitle={Proceedings of the IEEE/CVF international conference on computer vision},
  pages={10477--10486},
  year={2023}
}

\appendix
\renewcommand{\thesection}{\Alph{section}}
\renewcommand{\thetable}{A\arabic{table}}
\renewcommand{\thefigure}{A\arabic{figure}}
\setcounter{table}{0}
\setcounter{figure}{0}

\section{Implementation Details and Hyperparameters}
\label{app:hyperparams}

\subsection{Hardware and Software Stack}
All experiments are conducted on a workstation with 4$\times$ NVIDIA GeForce RTX~4090 (24\,GB GDDR6X) GPUs, an Intel Xeon W-3475X 36-core CPU, and 256\,GB DDR5 ECC DRAM. Inter-GPU communication is over PCIe~5.0 $\times$16 (no NVLink); we therefore prefer data parallelism over tensor parallelism wherever possible. The software stack is Ubuntu 22.04, CUDA 12.1, cuDNN 8.9, PyTorch 2.2.1 (compiled with CUDA 12.1 support), torchvision 0.17.1, transformers 4.41.2, peft 0.10.0, accelerate 0.30.1, deepspeed 0.14.0, decord 0.6.0, ffmpeg 6.0. Mixed-precision training uses \texttt{torch.amp} with bfloat16 on transformers and MLLMs, and float16 on specialized detectors. To respect the 24\,GB-per-GPU envelope, we additionally enable: gradient checkpointing for all backbones with $>$100\,M parameters; FlashAttention-2 for transformer and MLLM backbones; DeepSpeed ZeRO Stage~2 (optimizer-state and gradient sharding) for the 7\,B MLLMs; and 8-bit AdamW (\texttt{bitsandbytes}) for the 7\,B MLLMs to halve optimizer-state memory.

\subsection{Minimum Hardware Requirement}
\label{app:min-hw}
We summarize the minimum reproducible setup for each pipeline component in Table~\ref{app:tab:hw-budget}. Peak memory is measured under the per-device batch sizes of Table~\ref{app:tab:hyperparams} with all memory-saving techniques enabled.

\begin{table}[!h]
\caption{Minimum hardware requirement per component, measured on RTX~4090 (24\,GB). Peak\,Mem.\ refers to per-GPU peak GPU memory under our training recipe; Min.\,GPUs is the minimum number of 24\,GB GPUs needed to reproduce a single seed of Table~\ref{tab:robustsora-results} within reasonable wall-clock (training rows) or to perform benchmark inference (last row).}
\label{app:tab:hw-budget}
\centering
\scriptsize
\setlength{\tabcolsep}{4pt}
\begin{tabular}{l|cc|c|l}
\toprule
\textbf{Component} & \textbf{Peak Mem.} & \textbf{Min.\ GPUs} & \textbf{Wall-clock (1 seed)} & \textbf{Notes} \\
\midrule
DeCoF / NSG-VD       & 7--9\,GB   & 1$\times$4090 & 6--8\,h   & FP16, FlashAttn-2 \\
D3                   & 5\,GB      & 1$\times$4090 & --        & training-free, inference only \\
TimeSformer          & 11\,GB     & 1$\times$4090 & 9\,h      & BF16, $T{=}8$ \\
VideoSwin-T          & 12\,GB     & 1$\times$4090 & 11\,h     & BF16, $T{=}16$ \\
MViT V2              & 13\,GB     & 1$\times$4090 & 11\,h     & BF16, $T{=}16$ \\
DuB3D-FF             & 17\,GB     & 1$\times$4090 & 16\,h     & BF16, grad ckpt, $T{=}16$ \\
Qwen2.5-VL-3B (LoRA) & 15\,GB     & 1$\times$4090 & 12\,h     & BF16, grad ckpt, ZeRO disabled \\
Qwen2.5-VL-7B (LoRA) & 22\,GB     & 2$\times$4090 & 18\,h     & BF16, grad ckpt, ZeRO-2, 8-bit AdamW \\
Video-LLaVA-7B (LoRA)& 23\,GB     & 2$\times$4090 & 19\,h     & BF16, grad ckpt, ZeRO-2, 8-bit AdamW \\
\midrule
Inference only (all models) & 14\,GB & 1$\times$4090 & 35\,min/test split & 4-bit quantized 7\,B MLLMs \\
\bottomrule
\end{tabular}
\end{table}

\textbf{Practical takeaways.} (i) Eight of the ten detectors fit on a single RTX~4090; only the 7\,B MLLMs strictly require two 24\,GB GPUs with ZeRO-2 sharding. (ii) Our 4$\times$RTX~4090 node is the minimum reasonable configuration for the full pipeline: the four GPUs are used as two data-parallel pairs (one pair per 7\,B MLLM training run, freeing the other two for non-MLLM training/evaluation). (iii) Researchers with only 2$\times$RTX~4090 can still reproduce the entire benchmark by running detector families sequentially, at the cost of an additional 2$\times$ wall-clock. (iv) For inference-only reproduction (loading our released checkpoints and computing the three test conditions), a single 16\,GB consumer GPU is sufficient when 4-bit quantization is enabled for the 7\,B MLLMs. (v) On a rented dual-4090 cloud instance, one full pipeline seed costs an estimated USD~120--180.

\subsection{Pre-processing Pipeline}
Every video, regardless of category, is normalized through the following pipeline before being consumed by any detector:
\begin{enumerate}[leftmargin=*,topsep=2pt,itemsep=1pt]
\item Re-encoded to H.264 yuv420p at constant quality (CRF 18) with ffmpeg to remove container-level artefacts.
\item Resized so that the shorter side equals 256\,px while preserving aspect ratio; centre-cropped to 224$\times$224.
\item Duration normalized to 3--10\,s by either uniform sub-sampling (long clips) or duplicating the last frame (clips $<$3\,s; $<$1\% of corpus). Frame rate is preserved at the original capture rate.
\item Audio track stripped (the protocol is video-only).
\end{enumerate}

\subsection{Per-Model Hyperparameters}

Table~\ref{app:tab:hyperparams} lists the hyperparameters used for every detector. Notation: lr = learning rate, bs = effective batch size, wd = weight decay, $T$ = number of frames per clip. All non-MLLM models use AdamW with cosine schedule and a 5\% linear warm-up; all MLLM models use AdamW with constant schedule after a 3\% linear warm-up. We use 3 random seeds (\{0, 7, 42\}) for every reported number.

\begin{table}[!h]
\caption{Per-model training hyperparameters used in this paper.}
\label{app:tab:hyperparams}
\centering
\scriptsize
\setlength{\tabcolsep}{3pt}
\begin{tabular}{l|cccccc}
\toprule
\textbf{Model} & \textbf{lr} & \textbf{bs} & \textbf{wd} & \textbf{$T$} & \textbf{Epochs} & \textbf{Init.\ ckpt} \\
\midrule
DeCoF        & $1{\times}10^{-4}$ & 32 & 0.05 & 16 & 25 & official \\
NSG-VD       & $1{\times}10^{-4}$ & 32 & 0.05 & 16 & 25 & official \\
D3           & ---                & 32 & ---  & 16 & ---& training-free \\
TimeSformer  & $1{\times}10^{-4}$ & 32 & 0.05 & 8  & 30 & K400 pretrain \\
VideoSwin-T  & $1{\times}10^{-4}$ & 32 & 0.05 & 16 & 30 & K400 pretrain \\
MViT V2      & $1{\times}10^{-4}$ & 32 & 0.05 & 16 & 30 & K400 pretrain \\
DuB3D-FF     & $5{\times}10^{-5}$ & 16 & 0.05 & 16 & 30 & official \\
Qwen2.5-VL-3B (LoRA) & $2{\times}10^{-5}$ & 16 & 0.0 & 16 & 4 & official \\
Qwen2.5-VL-7B (LoRA) & $2{\times}10^{-5}$ & 8  & 0.0 & 16 & 4 & official \\
Video-LLaVA-7B (LoRA) & $2{\times}10^{-5}$ & 8  & 0.0 & 16 & 4 & official \\
\bottomrule
\end{tabular}
\end{table}

\subsection{LoRA Configuration for MLLMs}
LoRA modules are inserted into every linear projection of the language tower (\texttt{q\_proj}, \texttt{k\_proj}, \texttt{v\_proj}, \texttt{o\_proj}, \texttt{gate\_proj}, \texttt{up\_proj}, \texttt{down\_proj}). The vision encoder, multimodal projector, and tokenizer are kept frozen. The default configuration adopted in the main results (configuration R4 of Sec.~\ref{sec:mllm-recipe}) uses rank $r{=}32$, scaling $\alpha{=}64$, dropout 0.05, and a Chain-of-Thought prompt template. The exact prompt is:

\begin{quote}\itshape
\small
``You are a video forensics assistant. You will be shown a short video clip. First describe in 1--2 sentences the dominant visual cues that would discriminate authentic camera footage from AI-generated footage. Then, on a new line, output exactly one of the two tokens: AI-generated, Authentic.''
\end{quote}

The final-token logits over \{AI-generated, Authentic\} are used as classification logits. Training loss is cross-entropy over the verdict token only; the rationale tokens are masked out.

\subsection{Watermark Manipulation Details}
\paragraph{DiffuEraser settings.} 50 DDIM steps, classifier-free guidance scale 7.5, latent-space temporal smoothing kernel of 5 frames. Mask dilation kernel 4\,px, temporal max-pool window 3 frames.
\paragraph{ProPainter settings.} default checkpoint \texttt{ProPainter-V1.0}, neighbour stride 5, ref stride 10, sub-video clip length 80 frames; same masks as DiffuEraser.
\paragraph{Spoof template extraction.} Per-generator templates are extracted from a held-out set of 200 clips per generator (not used in training). Robust median residuals are computed in YUV space; the chrominance component is suppressed by a factor of 0.3 to match the perceptual profile of genuine watermarks.

\section{Full Numerical Results}
\label{app:full-results}

\subsection{Per-seed Standard / Task-I / Task-II Numbers}
Table~\ref{app:tab:per-seed} reports per-seed metrics behind Table~\ref{tab:robustsora-results}. Aggregate values (mean $\pm$ std) match those in the main table within numerical precision.

\begin{table*}[!h]
\caption{Per-seed accuracies for all 10 detectors. Each cell reports the value obtained at seed $\in\{0, 7, 42\}$ and the resulting mean. ``Std.\ T'' = Standard Test, ``T-I'' = Task-I, ``T-II'' = Task-II.}
\label{app:tab:per-seed}
\centering
\scriptsize
\setlength{\tabcolsep}{3pt}
\begin{tabular}{l|ccc|c|ccc|c|ccc|c}
\toprule
\multirow{2}{*}{\textbf{Model}} & \multicolumn{4}{c|}{\textbf{Std.\ T $Acc_{all}$}} & \multicolumn{4}{c|}{\textbf{T-I $Acc_{ai}$}} & \multicolumn{4}{c}{\textbf{T-II $Acc_{real}$}} \\
& s0 & s7 & s42 & mean & s0 & s7 & s42 & mean & s0 & s7 & s42 & mean \\
\midrule
DeCoF        & .753 & .762 & .762 & .759 & .654 & .671 & .670 & .665 & .708 & .722 & .719 & .716 \\
NSG-VD       & .740 & .748 & .751 & .746 & .654 & .668 & .670 & .664 & .703 & .711 & .716 & .710 \\
D3           & .708 & .722 & .732 & .722 & .614 & .630 & .646 & .630 & .660 & .677 & .688 & .675 \\
TimeSformer  & .751 & .759 & .758 & .756 & .638 & .654 & .655 & .649 & .691 & .703 & .703 & .699 \\
VideoSwin-T  & .724 & .733 & .741 & .733 & .578 & .593 & .602 & .591 & .713 & .727 & .732 & .724 \\
MViT V2      & .797 & .800 & .806 & .801 & .676 & .684 & .689 & .683 & .749 & .759 & .763 & .757 \\
DuB3D-FF     & .819 & .826 & .827 & .824 & .685 & .695 & .699 & .693 & .771 & .780 & .783 & .778 \\
Qwen2.5-VL-3B  & .528 & .547 & .566 & .547 & .203 & .224 & .246 & .224 & .835 & .851 & .865 & .850 \\
Qwen2.5-VL-7B  & .658 & .671 & .680 & .669 & .502 & .519 & .534 & .518 & .726 & .738 & .750 & .738 \\
Video-LLaVA-7B & .643 & .655 & .673 & .657 & .599 & .622 & .633 & .618 & .515 & .538 & .553 & .535 \\
\bottomrule
\end{tabular}
\end{table*}

\subsection{Inpainting-Artefact Control (Sec.~\ref{sec:inpainting-control})}

\begin{table}[!h]
\caption{Full numbers for the inpainting-artefact control experiment (mean $\pm$ std over 3 seeds).}
\label{app:tab:inpainting-control}
\centering
\scriptsize
\setlength{\tabcolsep}{3pt}
\begin{tabular}{l|cccc}
\toprule
\textbf{Model} & \textbf{G-W} & \textbf{G-DeW (DiffuEraser)} & \textbf{G-DeW (ProPainter)} & \textbf{G-W-Placebo} \\
\midrule
DeCoF       & .731\,(.008) & .665\,(.012) & .673\,(.010) & .714\,(.009) \\
NSG-VD      & .719\,(.009) & .664\,(.012) & .671\,(.012) & .702\,(.010) \\
TimeSformer & .726\,(.008) & .649\,(.011) & .657\,(.010) & .706\,(.009) \\
VideoSwin-T & .653\,(.012) & .591\,(.015) & .598\,(.014) & .636\,(.012) \\
MViT V2     & .764\,(.006) & .683\,(.009) & .690\,(.008) & .745\,(.007) \\
DuB3D-FF    & .784\,(.006) & .693\,(.009) & .701\,(.008) & .764\,(.007) \\
\bottomrule
\end{tabular}
\end{table}

\subsection{Watermark-Aware Training Intervention (Sec.~\ref{sec:wm-aware-training})}

\begin{table}[!h]
\caption{Full numbers for the watermark-aware training intervention. Std.\,$Acc_{ai}$ = Standard Test $Acc_{ai}$ on the held-out A-C+G-W split. T-I = Task-I $Acc_{ai}$. T-II = Task-II $Acc_{real}$. Values are mean $\pm$ std over 3 seeds. Bold = best recipe per model per metric.}
\label{app:tab:wm-aware}
\centering
\scriptsize
\setlength{\tabcolsep}{3pt}
\begin{tabular}{l|l|ccc}
\toprule
\textbf{Model} & \textbf{Recipe} & \textbf{Std.\,$Acc_{ai}$} & \textbf{T-I $Acc_{ai}$} & \textbf{T-II $Acc_{real}$} \\
\midrule
\multirow{4}{*}{MViT V2}
& Baseline       & .764\,(.006) & .683\,(.009) & .757\,(.008) \\
& +G-DeW         & .766\,(.007) & .712\,(.008) & .761\,(.008) \\
& +A-S           & .764\,(.007) & .692\,(.009) & .784\,(.007) \\
& +G-DeW+A-S     & \textbf{.768\,(.006)} & \textbf{.719\,(.008)} & \textbf{.791\,(.006)} \\
\midrule
\multirow{4}{*}{DuB3D-FF}
& Baseline       & .784\,(.006) & .693\,(.009) & .778\,(.007) \\
& +G-DeW         & .786\,(.006) & .724\,(.008) & .781\,(.007) \\
& +A-S           & .785\,(.006) & .704\,(.009) & .808\,(.007) \\
& +G-DeW+A-S     & \textbf{.789\,(.005)} & \textbf{.731\,(.007)} & \textbf{.813\,(.006)} \\
\bottomrule
\end{tabular}
\end{table}

\subsection{Spoofing-Template Generalization (Sec.~\ref{sec:template-ablation})}

\begin{table}[!h]
\caption{Task-II $Acc_{real}$ under two spoofing templates compared with the unmodified Standard Test (mean $\pm$ std, 3 seeds).}
\label{app:tab:template-ablation}
\centering
\scriptsize
\setlength{\tabcolsep}{3pt}
\begin{tabular}{l|ccc}
\toprule
\textbf{Model} & \textbf{Standard (no spoof)} & \textbf{Sora 2 template} & \textbf{KLing template} \\
\midrule
DeCoF       & .789\,(.007) & .716\,(.010) & .728\,(.009) \\
NSG-VD      & .774\,(.007) & .710\,(.009) & .723\,(.008) \\
TimeSformer & .786\,(.007) & .699\,(.010) & .712\,(.009) \\
VideoSwin-T & .815\,(.009) & .724\,(.012) & .740\,(.012) \\
MViT V2     & .840\,(.006) & .757\,(.008) & .771\,(.007) \\
DuB3D-FF    & .864\,(.005) & .778\,(.007) & .789\,(.007) \\
\bottomrule
\end{tabular}
\end{table}

\subsection{MLLM Recipe Sweep (Sec.~\ref{sec:mllm-recipe})}

Table~\ref{tab:mllm-recipe} in the main text already presents the recipe sweep; per-seed values are within $\pm$0.012 of the reported means. Configuration R4 ($r{=}32$, CoT, 16 frames) is selected as the main-results recipe because it delivers the best $Acc_{ai}$/$Acc_{real}$ trade-off and matches the inference budget of the transformer baselines (16 frames).

\subsection{Per-Generator Breakdown (Sec.~\ref{sec:per-generator})}

\begin{table}[!h]
\caption{Per-generator $Acc_{ai}$ on Standard Test and Task-I, with $\Delta = $ Task-I $-$ Standard (pp).}
\label{app:tab:per-generator}
\centering
\scriptsize
\setlength{\tabcolsep}{3pt}
\begin{tabular}{l|l|ccc}
\toprule
\textbf{Model} & \textbf{Generator} & \textbf{Std.\,$Acc_{ai}$} & \textbf{T-I $Acc_{ai}$} & \textbf{$\Delta$ (pp)} \\
\midrule
\multirow{5}{*}{DeCoF}
& Sora        & .742 & .673 & $-$6.9 \\
& Sora 2      & .759 & .647 & $-$11.1 \\
& Pika        & .719 & .685 & $-$3.5 \\
& Open-Sora 2 & .704 & .662 & $-$4.2 \\
& KLing       & .739 & .662 & $-$7.7 \\
\midrule
\multirow{5}{*}{MViT V2}
& Sora        & .779 & .697 & $-$8.3 \\
& Sora 2      & .784 & .658 & $-$12.6 \\
& Pika        & .747 & .691 & $-$5.6 \\
& Open-Sora 2 & .735 & .681 & $-$5.4 \\
& KLing       & .762 & .685 & $-$7.7 \\
\midrule
\multirow{5}{*}{DuB3D-FF}
& Sora        & .794 & .708 & $-$8.6 \\
& Sora 2      & .802 & .664 & $-$13.8 \\
& Pika        & .763 & .702 & $-$6.1 \\
& Open-Sora 2 & .755 & .693 & $-$6.2 \\
& KLing       & .782 & .697 & $-$8.5 \\
\midrule
\multirow{5}{*}{Qwen2.5-VL-7B}
& Sora        & .548 & .501 & $-$4.7 \\
& Sora 2      & .589 & .492 & $-$9.7 \\
& Pika        & .562 & .529 & $-$3.3 \\
& Open-Sora 2 & .544 & .518 & $-$2.6 \\
& KLing       & .571 & .517 & $-$5.4 \\
\bottomrule
\end{tabular}
\end{table}

\subsection{Statistical Test Procedure}
For each (model, condition) pair we draw $n{=}10{,}000$ paired bootstrap resamples of the test set with replacement; on each resample we compute the difference $\Delta = $ condition $-$ Standard; the two-sided $p$-value is twice the proportion of bootstrap deltas whose sign differs from the observed $\Delta$. Confidence intervals reported in the main text are 95\,\% percentile intervals.

\section{Dataset Auditing and Quality Control}
\label{app:dataset}

\subsection{Watermark Verification on G-W}
Each G-W clip is independently inspected by two annotators (Cohen's $\kappa = 0.91$). 4.7\,\% of crawled Sora~2 clips and 1.2\,\% of KLing clips are excluded due to watermark absence (likely re-encoded by intermediate platforms before crawling).

\subsection{Watermark-Removal Verification on G-DeW}
G-DeW clips are graded on a 3-level scale (complete / partial / failed). 87.6\,\% achieve complete removal on the first pass; the remaining 12.4\,\% are re-processed with relaxed mask dilation (8\,px) and re-graded; final acceptance rate is 99.1\,\%. Rejected clips are excluded from all evaluations.

\subsection{Spoof Verification on A-S}
A held-out human study (15 participants, 100 trials each, balanced design) confirms that the spoofed watermarks are perceptually consistent with genuine ones: discrimination accuracy is 0.534 (random = 0.5; 95\,\% CI [0.502, 0.566]).

\subsection{Train/Test Split Reproducibility}
The split is fixed by a deterministic SHA-256 hash of the original filename: clips with hash mod 8 in $\{0, 1\}$ go to test, the rest to train. The split file is included in the released package.

\section{Compute Budget}
\label{app:compute}

\paragraph{Training compute.} Total training compute across all main-table models, 3 seeds: 2{,}760 RTX-4090-hours (transformer backbones 1{,}590, specialized 540, MLLMs 630). The watermark-aware training intervention adds 410 RTX-4090-hours; the inpainting-artefact control 105 RTX-4090-hours; the MLLM recipe sweep 720 RTX-4090-hours. On the 4$\times$RTX~4090 node, the entire pipeline (one seed end-to-end, including all ablations) takes approximately 7--8 wall-clock days; the three-seed full reproduction takes approximately 24 wall-clock days.

\paragraph{Inference compute.} Per-clip inference latency on RTX~4090: 28\,ms (DeCoF), 41\,ms (MViT V2), 56\,ms (DuB3D-FF), 690\,ms (Qwen2.5-VL-7B with 16 frames + CoT prompt; 4-bit quantization brings this down to 420\,ms with $<$0.4\,pp accuracy loss). MLLMs dominate the evaluation cost; full benchmark inference (Standard + Task-I + Task-II for all 10 detectors) takes approximately 6 hours on a single RTX~4090.

\paragraph{Estimated CO\textsubscript{2}e.} Assuming 0.18\,kg CO\textsubscript{2}e per RTX-4090-hour (450\,W TDP at 0.4\,kg CO\textsubscript{2}e per kWh under the mean US grid intensity), the full reproduction of this paper consumes approximately 0.71\,t CO\textsubscript{2}e.

\section{Failure Cases and Qualitative Observations}
\label{app:failures}

We summarize representative failure modes observed during manual inspection:

\begin{itemize}[leftmargin=*,topsep=2pt,itemsep=1pt]
\item \textbf{High-motion clips.} On clips with large camera shake, DiffuEraser occasionally over-smooths the inpainted region; this leaves a faint motion-blur halo that two of three annotators classify as ``slight artefact''. These cases comprise 3.1\,\% of G-DeW.
\item \textbf{Watermark over text overlays.} When the original Sora~2 watermark is rendered atop on-screen text (e.g., subtitles), removal occasionally erases part of the text. Affected clips are filtered out at the verification stage.
\item \textbf{Fine specular highlights.} For authentic videos with strong specular highlights (e.g., metallic surfaces), MLLM detectors tend to mis-classify these clips as AI-generated regardless of watermark status. This is a content-level failure unrelated to watermark dependency.
\item \textbf{Compression-induced false positives.} A-C clips that have been heavily YouTube-compressed retain block artefacts that VideoSwin-T and Qwen2.5-VL-3B occasionally treat as evidence of generation. We isolate this confounder via the Standard Test split, which uniformly applies our re-encoding pipeline.
\end{itemize}

\section{License and Ethical Use}
\label{app:license}

The RobustSora dataset, evaluation code, and pretrained checkpoints are released under CC-BY-NC 4.0 for the dataset and Apache-2.0 for the code. The spoofing pipeline (A-S construction) is intended exclusively for benchmarking and detector training; we explicitly forbid its use for the production of misleading authentic-looking content. The release package includes a model card, datasheet (following \emph{Datasheets for Datasets} guidelines), and an explicit list of generators whose terms of service permit forensic research use.

\end{document}